\documentclass{article}

\usepackage[preprint]{neurips_2026}

\usepackage[utf8]{inputenc} 
\usepackage[T1]{fontenc}    
\usepackage{hyperref}       
\usepackage{url}            
\usepackage{booktabs}       
\usepackage{amsfonts}       
\usepackage{nicefrac}       
\usepackage{microtype}      
\usepackage{xcolor}         
\usepackage{amsmath}
\usepackage{tikz}
\usepackage{multirow}
\usepackage{algorithm}
\usepackage{algpseudocode}
\usetikzlibrary{arrows.meta, positioning, calc}

\title{ConTex: Reformulating Counterfactual Generation For Time Series Forecasting}

\author{
  Jan Voets \and Hasan Tercan \and Tobias Meisen \and Sebastian Baum \\
  Institute for Technologies and Management of Digital Transformation\\
  University of Wuppertal\\
  \texttt{voets@uni-wuppertal.de}
}

\begin{document}

\maketitle

\begin{abstract}
Decision-making with deep learning–based time series forecasting requires not only accurate predictions but also actionable insights. However, current architectures do not inherently provide such information, which calls for methodologies that can extract relevant insights and explain model behavior. Specifically, in the context of decision-making, guidance is needed on how current conditions must be modified to shift from a predicted outcome to a desired future scenario. Counterfactual explanations provide a natural framework for this task, as they represent minimal input changes that alter the model’s prediction, supporting decision-making by indicating when and how intervention is required. However, existing approaches rely on instance-wise optimization, leading to inconsistency across different instances, high computational costs, and thus limited applicability in real-time settings. To address these limitations, we reformulate counterfactual generation for time series forecasting as the problem of learning a globally consistent intervention strategy, allowing counterfactuals to be generated for any input sample through a single, shared function. We propose Counterfactual Time Series Explanations (ConTex), a model-agnostic, decomposed architecture comprising a temporal context encoder and a conditional encoder, followed by two heads that capture interventions in terms of temporal relevance and modification strength. This structure overcomes the instability and inconsistency of instance-based counterfactual generation by producing counterfactuals through targeted, interpretable interventions across time and feature dimensions in a single forward pass, making it suitable for real-time applications. We demonstrate this across multiple forecasting architectures and benchmark datasets, achieving state-of-the-art validity while generating sparse counterfactuals that minimize the number of necessary interventions. Additionally, our approach achieves an overall computational reduction of at least 12–36× compared to instance-wise generation and supports real-time inference at approximately 0.007 seconds inference time. 
\end{abstract}

\section{Introduction}
Time series forecasting plays a critical role in a wide range of real-world applications, including energy systems, finance, weather prediction, healthcare, and supply chain management. While modern forecasting models achieve strong predictive performance, they remain largely opaque, limiting their usability in high-stakes decision-making scenarios \cite{kim_comprehensive_2025}. For example, in energy systems, operators may wish to understand how current demand or generation patterns must be adjusted to avoid future peak loads, which is crucial for maintaining grid stability and enabling peak-shaving strategies \cite{AMARAOUALI20231272, dai2021electricalpeakdemandforecasting}. Meanwhile, in healthcare, clinicians may seek to identify how a patient's current state must change to prevent predicted deterioration from historic patient data \cite{ko_deep_2023}.

Explanation methods for time-series forecasting are typically retrospective: given an observed outcome, they aim to identify the factors that led to it, for instance, through feature attribution methods (e.g., TimeSHAP \cite{Bento_2021}, WindowSHAP \cite{nayebi2023windowshapefficientframeworkexplaining}, TSHAP \cite{10.1007/978-3-032-06078-5_4}), attention-based interpretations in transformer models (e.g., Temporal Fusion Transformers \cite{lim2020temporalfusiontransformersinterpretable}, PatchTST \cite{nie2023timeseriesworth64}), or a broad class of post-hoc explanation techniques \cite{theissler2022explainable}, as well as interpretable representation-based approaches (e.g., EXCODER \cite{hahn2026excoderexplainableclassificationdiscrete}). In contrast, decision-making requires a forward-looking perspective, whereby one must understand how to alter present conditions to achieve a desired future outcome. This reveals a fundamental gap: while forecasting models implicitly encode how present conditions shape future trajectories, they do not provide actionable guidance on how they could be understood and leveraged. For continuous time series data, the space of possible modifications is effectively unbounded and affects both the time and feature dimensions. For guidance to be actionable, any modifications should be minimal. In other words, only the necessary changes should be introduced to keep the time series as close as possible to its original form. The resulting alternative series of such minimum modification is referred to as a counterfactual \cite{wachter2018counterfactualexplanationsopeningblack}.

Existing counterfactual generation methods for time series forecasting suffer from several limitations. Most approaches rely on instance-wise optimization, solving a separate problem for each sample \cite{wang_counterfactual_2023, luo_counterfactual_2026, wang_comet_2024}, which leads to inconsistent, hard-to-interpret modifications and high computational costs, limiting applicability in real-time or interactive settings. More generally, generating counterfactuals for time series poses challenges in maintaining temporal coherence, plausibility, and sparsity of interventions \cite{schlegel2026whatifexplanationstimecounterfactuals}. Additionally, existing methods often use target trajectories that depend on the input. For example, Wang et al. used input-dependent bounds or trend adjustments \cite{wang_counterfactual_2023}, which improve feasibility by defining easy-to-satisfy targets but couple the objective to the original time series. This shifts part of the burden from modifying the input to adapting the objective itself, which obscures the true difficulty of achieving the desired outcome. Consequently, it fails to address the question of how the present must be altered to achieve a predefined future scenario.

To overcome these limitations, we propose a different generation paradigm for counterfactual explanations in time series forecasting. Instead of solving a separate optimization problem for each instance, we learn a global intervention function that maps input states and target conditions to effective interventions. This amortized approach generates counterfactuals in a single forward pass and enables consistent intervention behavior across instances. Since interventions in time series are inherently structured along both temporal and feature dimensions, we therefore introduce ConTex, a neural architecture that decomposes them into temporal relevance and modification strength. This decomposition yields interpretable and structured counterfactuals by indicating where and how the input should be modified. In contrast to optimization-based approaches, ConTex learns generalizable intervention patterns and transforms any given time series into its counterfactual in a single forward pass at inference time. In summary, our contributions are as follows:

\begin{itemize}

\item We shift counterfactual generation from instance-wise optimization to amortized intervention learning. This function maps input states and target conditions to effective interventions, enabling counterfactuals to be generated in a single forward pass.

\item We propose ConTex, an architecture that parameterizes interventions by two components: a temporal relevance mask that indicates where to modify and a modification strength that specifies how to modify. This leads to counterfactuals that are sparse, local, and interpretable.

\item We introduce target-conditioned attribution for time series forecasting, identifying which timesteps must be modified to achieve a desired outcome. This yields a forward-looking, intervention-based temporal relevance representation, highlighting actionable timesteps and providing an alternative to retrospective post-hoc attribution methods. We visualize this as a heatmap indicating the likelihood of achieving the desired result through modification.

\item By evaluating and comparing ConTex on four forecasting backbones (PatchTST, N-HiTS, DLinear, TiDE) and four benchmark datasets (NN5, Electricity, Tourism, M4), compared to per instance optimization, ConTex achieves the best validity in 16 of 16 cases and the best compactness in 14 of 16 cases, while reducing the generation cost by 12–36 times end-to-end.

\end{itemize}

\section{Related Work}
Counterfactual explanations were introduced by \cite{wachter2018counterfactualexplanationsopeningblack} as minimal input perturbations that change a model's prediction. In time series, counterfactual generation spans optimization-based, instance-based, and generative latent-space approaches \cite{schlegel2026whatifexplanationstimecounterfactuals}. Further, existing work has predominantly focused on classification settings rather than forecasting (e.g., \cite{inproceedings111, delaney2021instancebasedcounterfactualexplanationstime, bahri2022shapeletbasedcounterfactualexplanationsmultivariate, yan2023selfinterpretabletimeseriesprediction}).

In the context of time series forecasting, particularly for actionable interpretability, most existing work focuses on scenario exploration rather than explicit counterfactual generation. Probabilistic models such as DeepAR \cite{SALINAS20201181} generate multiple plausible future trajectories, while architectures like the Temporal Fusion Transformer \cite{lim2020temporalfusiontransformersinterpretable} enable scenario analysis by conditioning on known future covariates. More closely related are approaches that incorporate causal reasoning into forecasting models \cite{ferchichi_trustworthy_2025}, enabling “what-if” analyzes. However, these methods are embedded within the predictive model and do not aim to compute minimal input interventions for a fixed predictor.

For counterfactual generation in time series forecasting, ForecastCF \cite{wang_counterfactual_2023} is among the first methods to explicitly address this problem by iteratively perturbing inputs to achieve desired trend changes. Subsequent works extend this optimization-based paradigm to domains such as cryptocurrency forecasting \cite{luo_counterfactual_2026} and multivariate settings \cite{wang_comet_2024}. A related line of work focuses on generating plausible future trajectories rather than minimal interventions. For example, \cite{zuin_navigating_2024} model future time series distributions via quantile regression and search for trajectories satisfying desired outcome constraints. While this improves statistical plausibility, the approach does not explicitly optimize minimal input changes for a fixed predictive model, distinguishing it from counterfactual formulations.

In contrast to existing approaches, our method is neither integrated into the forecasting model nor based on instance-wise optimization. Instead, we replace instance-wise optimization by learning a global intervention function that predicts input modifications in a single forward pass, enabling efficient, real-time generation of counterfactuals across instances derived from a global policy. By decomposing interventions into temporal relevance and modification strength, our approach provides interpretable insights into both where and how the input must be changed. However, unlike plausibility-based approaches (e.g. \cite{zuin_navigating_2024}), ConTex does not explicitly enforce distributional constraints, instead relying on implicit regularization through temporal context modeling and minimal-intervention objectives.

\section{Problem Formulation}

We define a desired forecast as a trajectory within a set of bounds:
\begin{equation}
\mathcal{Y}_{\text{target}}
=
\left\{
y \in \mathbb{R}^{H}
\;\middle|\;
L_t \le y_t \le U_t,\;
L_t \le U_t \;\forall t \in \{1,\dots,H\}
\right\}
\end{equation}
where $L \in \mathbb{R}^{H}$ and $U \in \mathbb{R}^{H}$ denote the lower and upper target bounds over horizon $H$.

Given an univariate time series $x \in \mathbb{R}^{T_{\mathrm{in}}}$, where $T_{\mathrm{in}}$ denotes the input sequence length, and a forecasting model $f_\theta : \mathbb{R}^{T_{\mathrm{in}}} \rightarrow \mathbb{R}^{H}$ with fixed parameters, the counterfactual problem consists of finding a minimally modified input $x'$ such that the resulting forecast satisfies the target constraint while preserving as much of $x$ as possible:
\begin{equation}
\label{eq:counterfactual_problem}
x^{\mathrm{cf}}
=
\arg\min_{x'} \mathcal{R}(x',x)
\quad \text{s.t.} \quad
f_\theta(x') \in \mathcal{Y}_{\text{target}}.
\end{equation}

Here, preservation is measured by $\mathcal{R}(x',x) = \|x' - x\|_2 + \frac{1}{T_{\mathrm{in}}} \sum_{t=1}^{T_{\mathrm{in}}} \mathbf{1}\!\left[|x_t - x'_t| > \tau \right]$
where the first term penalizes scale of deviations from the original input, while the second penalizes the number of altered timesteps. This formulation can be interpreted as a \emph{local constrained inverse problem}: rather than inverting $f_\theta$ globally, we seek an input $x'$ in the vicinity of $x$ whose prediction $f_\theta(x')$ satisfies the desired $\mathcal{Y}_{\text{target}}$.

We further simplify this problem by expressing the counterfactual as an additive modification of the original input,
\(
x^{\mathrm{cf}} = x + z,
\)
where $z \in \mathbb{R}^{T_{\mathrm{in}}}$ denotes a minimal intervention. This allows us to reformulate the problem as
\begin{equation}
\label{eq:intervention_problem}
z^* = \arg\min_{z} \mathcal{R}(x + z, x)
\quad \text{s.t.} \quad
f_\theta(x + z) \in \mathcal{Y}_{\text{target}}.
\end{equation}

Instead of directly optimizing over $z$, we seek an interpretable and sparse solution, and accordingly parameterize the intervention in a structured manner by decomposing it into temporal components. Specifically, we define a function that predicts (i) whether a modification is required and (ii) how strongly it should be applied:
\begin{align}
\label{eq:intervention_decomposition}
(m, s) &= G_\theta(x, \hat{y}, c), \\
z &= m \odot s,
\end{align}
where $G_\theta$ predicts a temporal relevance mask 
$m \in [0,1]^{T_{\mathrm{in}}}$  and modification strength 
$s \in \mathbb{R}^{T_{\mathrm{in}}}$, conditioned on the input $x$, 
its current prediction $\hat{y} = f(x)$, and conditioning vector 
$c$, a compact representation of $\mathcal{Y}_{\text{target}}$.

\section{Method}
\subsection{ConTex Architecture}

\begin{figure}[!htbp]
\small
\centering
\resizebox{0.95\columnwidth}{!}{%
\begin{tikzpicture}[
    font=\footnotesize,
    >=Latex,
    node distance=0.45cm and 0.65cm,
    box/.style={
        draw,
        rounded corners,
        minimum width=1.55cm,
        minimum height=0.55cm,
        align=center
    },
    smallbox/.style={
        draw,
        minimum width=0.55cm,
        minimum height=0.38cm,
        align=center
    },
    op/.style={
        draw,
        rounded corners,
        minimum width=1.35cm,
        minimum height=0.50cm,
        align=center
    },
    arr/.style={->, thick}
]

\node[smallbox] (x) {$x$};
\node[smallbox, below=0.45cm of x] (condin) {$\hat{y},\, c$};

\node[box, right=0.6cm of x] (tempenc) {Temporal\\Encoder};
\node[box, right=0.6cm of condin] (condenc) {Condition\\Encoder};

\node[box, right=1.5cm of $(tempenc)!0.5!(condenc)$] (film) {FiLM\\Mod.};

\node[box, right=0.9cm of film, yshift=0.35cm] (maskhead) {Mask\\Head};
\node[box, right=0.9cm of film, yshift=-0.35cm] (strengthhead) {Strength\\Head};

\node[smallbox, right=0.55cm of maskhead] (m) {$m$};
\node[smallbox, right=0.55cm of strengthhead] (s) {$s$};

\node[op, right=0.85cm of $(m)!0.5!(s)$] (combine) {$z=m\odot s$};
\node[op, right=0.75cm of combine] (xcf) {$x_{\mathrm{cf}}=x+z$};

\draw[arr] (x) -- (tempenc);
\draw[arr] (condin) -- (condenc);

\draw[arr] (tempenc.east) -- node[above] {$h_x$} (film.west);
\draw[arr] (condenc.east) -- node[below] {$(\gamma,\beta)$} (film.west);

\draw[arr] (film.east) -- node[above] {$\tilde{h}_x$} (maskhead.west);
\draw[arr] (film.east) -- node[below] {$\tilde{h}_x$} (strengthhead.west);

\draw[arr] (maskhead) -- (m);
\draw[arr] (strengthhead) -- (s);

\draw[arr] (m) -- (combine);
\draw[arr] (s) -- (combine);
\draw[arr] (combine) -- (xcf);

\end{tikzpicture}%
}
\caption{ConTex architecture. A target-conditioned model using feature-wise linear modulation (FiLM, \cite{perez2017filmvisualreasoninggeneral}) to adapt temporal features. The modulated representation is used to predict a temporal mask $m \in [0,1]^T$ and a signed modification strength $s \in \mathbb{R}^{T \times D}$. Their composition yields the intervention $z = m \odot s$, which is added to the input to obtain the counterfactual sequence.}
\label{fig:contex_architecture}
\end{figure}
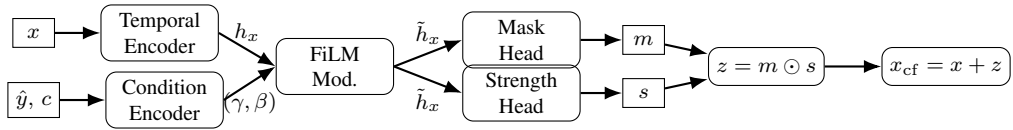

Addressing the formulated problem requires an architecture that can decide whether a given timestep should be modified based on multiple factors: (i) its temporal context, (ii) its position within the sequence, (iii) its feature value, and (iv) the desired target trajectory. These requirements introduce multiple interdependent sources of complexity. A naive approach would attempt to model all of these aspects within a single monolithic component. However, this quickly becomes problematic, as each of these subtasks is structurally distinct. We therefore follow a modular design principle, where each component is responsible for a single specified function. In particular, we separate (i) the extraction of temporal context $h_x$ from the input sequence and (ii) the encoding of the desired target condition. These representations are combined through feature-wise linear modulation $\tilde{h}_x = h_x \odot (1 + \gamma) + \beta$, enabling the model to adapt its behavior to different desired trajectories. Additionally, the condition embedding is tiled across time and concatenated to the temporal representation before the prediction heads. 

Since the input exhibits temporal dependencies, the temporal encoder must capture both local and long-range patterns, which can be achieved using either recurrent or convolutional architectures. We evaluate both a bidirectional LSTM (BiLSTM) \cite{10.1162/neco.1997.9.8.1735} and a temporal convolutional network (TCN) \cite{bai2018empiricalevaluationgenericconvolutional}, with the TCN yielding superior performance. This is consistent with our objective of producing sparse and interpretable modifications, as the TCN constructs representations based on fixed receptive fields, enabling more explicit identification of localized patterns compared to the aggregated representations obtained through sequential propagation in recurrent models. In contrast, the condition is not provided as a sequence but as a fixed-size vector. Accordingly, the condition encoder is implemented as a multilayer perceptron (MLP) that maps $c$, comprising the desired trajectory center, width, current forecast, and slope, into a latent embedding. For full implementation details, see Appendix~\ref{full_implementation}.

\subsection{Training Objective}
\label{sec:training_objective}

Counterfactual generation involves a trade-off between achieving the desired forecast behavior and minimizing the magnitude and extent of the intervention. We therefore optimize a composite objective consisting of four components.

\paragraph{Forecast alignment.} To reach high validity, we need to (i) ensure that a high percentage of timesteps satisfy the bounds and (ii) that the forecast follows accurately the desired trend. Accordingly we penalize deviations from the midpoint of the target bounds as well as the (smooth) fraction of timesteps violating the constraints. Let
\(
\mu_t = \frac{U_t + L_t}{2}, \quad
w_t = \max\!\left(\frac{U_t - L_t}{2}, \varepsilon\right), \quad
\delta_t = \frac{\hat y_t - \mu_t}{w_t}.
\)
\begin{equation}
\mathcal{L}_{\mathrm{center}} =
\frac{1}{H}\sum_{t=1}^H |\delta_t|,
\quad
\mathcal{L}_{\mathrm{valid}} =
\frac{1}{H}\sum_{t=1}^H
\sigma\!\left(\kappa (|\delta_t|-1)\right).
\end{equation}
Here, $\delta_{t,d}$ converts each constraint into a common coordinate system in which the boundary is at $\pm 1$, to enable $\mathcal{L}_{\mathrm{center}}$ to measure distance to the desired trajectory and $\mathcal{L}_{\mathrm{valid}}$ to quantify violation frequency consistently across all timesteps. The steepness of the sigmoid transition is controlled by $\kappa>0$.

\paragraph{Intervention regularization.}
To ensure minimal interventions, both the magnitude of changes and the number of modified timesteps must be controlled. Accordingly, we impose proximity and sparsity constraints addressing each objective:

\begin{equation}
\mathcal{L}_{\mathrm{prox}} =
\frac{1}{T_{\mathrm{in}}}
\sum_{t=1}^{T_{\mathrm{in}}}
\left|x^{\mathrm{cf}}_{t} - x_{t}\right|,
\quad
\mathcal{L}_{\mathrm{sparse}} =
\frac{\|m\|_1}{\|m\|_2}
+
\frac{1}{T_{\mathrm{in}}}
\sum_{t=1}^{T_{\mathrm{in}}} m_t(1-m_t).
\end{equation}
where $T_{\mathrm{in}}$ denotes the input sequence length. To avoid over-penalizing localized but necessary deviations, we use an $L_1$ proximity loss instead of $L_2$, promoting sparse and interpretable interventions over weak distributional shifting. Furthermore, to avoid degenerate solutions with uniformly distributed masks, $\mathcal{L}_{\mathrm{sparse}}$ combines two terms: the first promotes compact interventions by encouraging concentrated mask activations, while the second acts as a binarization regularizer, pushing the mask towards near-binary decisions.
\paragraph{Total loss.}
The final objective is then given by
\[
\mathcal{L}_{\mathrm{total}} =
\lambda_{\mathrm{center}} \mathcal{L}_{\mathrm{center}} +
\lambda_{\mathrm{valid}} \mathcal{L}_{\mathrm{valid}} +
\lambda_{\mathrm{prox}} \mathcal{L}_{\mathrm{prox}} +
\lambda_{\mathrm{sparse}} \mathcal{L}_{\mathrm{sparse}},
\]
balancing target satisfaction and minimal intervention. As our objective is to obtain valid and sparse counterfactuals, we employ an early stopping criterion that prioritizes validity and compactness. Specifically, training is stopped if a combined metric of validity and sparsity does not improve.

\section{Experiments}

\subsection{Experimental Setup}
\label{Forecasting_datasets_and_model_selection}

\begin{table}[!htbp]
\centering
\footnotesize
\begin{tabular}{lccccc}
\toprule
Dataset & Resolution & Input Length & Horizon & Characteristics & Source \\
\midrule
M4          & Daily        & 42  & 14  & Smooth, trend-dominated & \cite{MAKRIDAKIS202054} \\
NN5         & Daily        & 168 & 56  & Noisy, long-horizon     & \cite{ANDRAWIS2011672} \\
Tourism     & Monthly      & 72  & 24  & Strong seasonality      & \cite{godahewa_tourism_2020} \\
Electricity & Hourly       & 288 & 96  & High-frequency          & \cite{godahewa_electricity_2020} \\
\bottomrule
\end{tabular}
\caption{Overview of datasets used in our experiments.}
\label{tab:datasets}
\end{table}

\paragraph{Datasets and forecasting models.} In order to evaluate the robustness and general applicability of the proposed method, we consider several datasets that span a diverse range of time series characteristics, as described in Table \ref{tab:datasets}. Similarly, we select forecasting models covering diverse architectural paradigms, including linear decomposition-based models (DLinear, \cite{zeng2022transformerseffectivetimeseries}), hierarchical MLP-based architectures (N-HiTS, \cite{challu2022nhitsneuralhierarchicalinterpolation}), encoder–decoder architectures (TiDE, \cite{das2024longtermforecastingtidetimeseries}), and transformer-based models (PatchTST, \cite{nie2023timeseriesworth64}).

\paragraph{Training protocol.} For every dataset, all forecasting models are independently optimized, frozen, and reused across all counterfactual methods to ensure consistency and fairness. The resulting forecasting performance is reported in the appendix, Table \ref{appendix:tab_smape_table}.  Architectural hyperparameters of ConTex are tuned once on the NN5 dataset, a challenging setting with long horizons and complex dynamics, and kept fixed across all other datasets.  Only sparsity and proximity weights, learning rate, and dropout are tuned per dataset via a lightweight search on the best-performing model. Full details are in \ref{appendix:hyperparameter}.

\paragraph{Baselines.}
We compare ConTex against three baselines covering different approaches to counterfactual generation. ForecastCF \cite{wang_counterfactual_2023} represents an instance-wise, perturbation-based generation. We further apply early stopping to ensure convergence. In addition, we include a heuristic baseline inspired by BaseShift \cite{wang_counterfactual_2023}. However, as our targets are defined by externally specified trajectories, we adapt BaseShift by applying an additive shift, aligning the input mean with the target level. Finally, we include an adapted version of BaseNN \cite{wang_counterfactual_2023}, which retrieves counterfactuals by selecting the nearest neighbor that satisfies the target constraint or maximizes validity with respect to the target set, given an input $x$. Further details can be found in \ref{appendix:basenn}. While not providing interventions, BaseNN approximates the highest validity attainable when relying solely on existing data.

\paragraph{Target Calibration.} Instance-specific or manually designed targets introduce bias and reduce comparability across datasets due to their lack of (i) consistent control over task difficulty, (ii) evaluation across the full dataset distribution, and (iii) assessment of behavioral consistency across datasets. In consequence, to ensure a fair and systematic evaluation, we adopt a controlled target family spanning the data distribution consistently across datasets. However, to avoid introducing artificial structure, we restrict this family to linear trajectories with tolerance, including constant and increasing trends, capturing fundamental interventions such as level shifts and trend adjustments while enabling domain-agnostic evaluation. Furthermore, the difficulty of generating a counterfactual for a target $\mathcal{Y}_{\text{target}}$ depends on (i) the width of the interval for $f(x+z)$ and (ii) its position in the forecast distribution, with low-density targets being harder to achieve. We therefore calibrate targets using percentile-based intervals of the model’s output distribution. As a reference, we set the interval such that a simple baseline attains $\sim 50\%$ validity (e.g. see Appendix \ref{dlinearcalibration}), representing moderate difficulty, and vary percentiles to control difficulty. Increasing trends are sampled from upper percentiles (e.g., $90^{\text{th}}$) of the slope distribution at fixed interval width, ensuring standardized and challenging targets across datasets.

\paragraph{Evaluation Metrics.} We evaluate counterfactual quality using validity and data-manifold closeness metrics following \cite{wang_counterfactual_2023}. Validity Ratio and Stepwise Validity AUC (S-AUC) measure target satisfaction and temporal consistency, while Proximity and Compactness quantify deviation from the original input. Full metric definitions are provided in Appendix~\ref{EvaluationMetrics}.

\subsection{Counterfactual Benchmark}

\begin{table}[!htbp]
\centering
\tiny
\setlength{\tabcolsep}{3pt}
\begin{tabular}{llcccccccc}
  \toprule
Model & CF Model & \multicolumn{2}{c}{NN5} & \multicolumn{2}{c}{Electricity} & \multicolumn{2}{c}{Tourism} & \multicolumn{2}{c}{M4} \\
  \midrule
 &  & Ratio $\uparrow$ & S-AUC $\uparrow$ & Ratio $\uparrow$ & S-AUC $\uparrow$ & Ratio $\uparrow$ & S-AUC $\uparrow$ & Ratio $\uparrow$ & S-AUC $\uparrow$ \\
  \midrule
\multirow{4}{*}{PatchTST} & ForecastCF & 0.324${\scriptstyle\pm0.160}$ & 0.079${\scriptstyle\pm0.056}$ & 0.705${\scriptstyle\pm0.210}$ & 0.384${\scriptstyle\pm0.261}$ & 0.482${\scriptstyle\pm0.225}$ & 0.365${\scriptstyle\pm0.215}$ & 0.707${\scriptstyle\pm0.114}$ & 0.689${\scriptstyle\pm0.114}$ \\
 & ConTex & 0.683${\scriptstyle\pm0.041}$ & 0.406${\scriptstyle\pm0.095}$ & \underline{\textbf{0.964${\scriptstyle\pm0.003}$}} & \underline{\textbf{0.911${\scriptstyle\pm0.007}$}} & 0.812${\scriptstyle\pm0.017}$ & 0.676${\scriptstyle\pm0.025}$ & 0.997${\scriptstyle\pm0.001}$ & 0.997${\scriptstyle\pm0.001}$ \\
 & BaseShift & 0.602 & 0.178 & 0.842 & 0.713 & 0.676 & 0.548 & 0.996 & 0.996 \\
 & BaseNN & \textbf{0.918} & \textbf{0.770} & 0.809 & 0.530 & \underline{\textbf{0.863}} & \underline{\textbf{0.754}} & \underline{\textbf{1.000}} & \underline{\textbf{1.000}} \\
  \midrule
\multirow{4}{*}{N-HiTS} & ForecastCF & 0.519${\scriptstyle\pm0.117}$ & 0.199${\scriptstyle\pm0.089}$ & 0.758${\scriptstyle\pm0.167}$ & 0.488${\scriptstyle\pm0.276}$ & 0.476${\scriptstyle\pm0.227}$ & 0.362${\scriptstyle\pm0.213}$ & 0.847${\scriptstyle\pm0.067}$ & 0.825${\scriptstyle\pm0.082}$ \\
 & ConTex & \textbf{0.971${\scriptstyle\pm0.026}$} & \textbf{0.921${\scriptstyle\pm0.062}$} & \textbf{0.941${\scriptstyle\pm0.005}$} & \textbf{0.851${\scriptstyle\pm0.008}$} & 0.764${\scriptstyle\pm0.048}$ & 0.656${\scriptstyle\pm0.065}$ & 0.997${\scriptstyle\pm0.000}$ & 0.996${\scriptstyle\pm0.001}$ \\
 & BaseShift & 0.648 & 0.208 & 0.858 & 0.720 & 0.681 & 0.536 & 0.996 & 0.995 \\
 & BaseNN & 0.914 & 0.689 & 0.859 & 0.603 & \textbf{0.851} & \textbf{0.737} & \underline{\textbf{1.000}} & \underline{\textbf{1.000}} \\
  \midrule
\multirow{4}{*}{DLinear} & ForecastCF & 0.398${\scriptstyle\pm0.140}$ & 0.118${\scriptstyle\pm0.048}$ & 0.620${\scriptstyle\pm0.270}$ & 0.295${\scriptstyle\pm0.245}$ & 0.381${\scriptstyle\pm0.212}$ & 0.266${\scriptstyle\pm0.180}$ & 0.836${\scriptstyle\pm0.082}$ & 0.791${\scriptstyle\pm0.095}$ \\
 & ConTex & \underline{\textbf{0.990${\scriptstyle\pm0.003}$}} & \underline{\textbf{0.949${\scriptstyle\pm0.003}$}} & \textbf{0.941${\scriptstyle\pm0.002}$} & \textbf{0.748${\scriptstyle\pm0.022}$} & \textbf{0.841${\scriptstyle\pm0.019}$} & \textbf{0.722${\scriptstyle\pm0.031}$} & 0.997${\scriptstyle\pm0.000}$ & 0.997${\scriptstyle\pm0.000}$ \\
 & BaseShift & 0.623 & 0.181 & 0.814 & 0.619 & 0.579 & 0.427 & 0.995 & 0.995 \\
 & BaseNN & 0.894 & 0.579 & 0.805 & 0.511 & 0.839 & 0.705 & \underline{\textbf{1.000}} & \underline{\textbf{1.000}} \\
  \midrule
\multirow{4}{*}{TiDE} & ForecastCF & 0.621${\scriptstyle\pm0.095}$ & 0.206${\scriptstyle\pm0.097}$ & 0.705${\scriptstyle\pm0.230}$ & 0.357${\scriptstyle\pm0.258}$ & 0.493${\scriptstyle\pm0.242}$ & 0.375${\scriptstyle\pm0.239}$ & 0.827${\scriptstyle\pm0.074}$ & 0.805${\scriptstyle\pm0.072}$ \\
 & ConTex & \textbf{0.976${\scriptstyle\pm0.006}$} & \textbf{0.884${\scriptstyle\pm0.034}$} & \textbf{0.940${\scriptstyle\pm0.002}$} & \textbf{0.755${\scriptstyle\pm0.013}$} & 0.773${\scriptstyle\pm0.018}$ & 0.679${\scriptstyle\pm0.018}$ & 0.970${\scriptstyle\pm0.038}$ & 0.968${\scriptstyle\pm0.043}$ \\
 & BaseShift & 0.654 & 0.201 & 0.838 & 0.699 & 0.675 & 0.548 & 0.995 & 0.995 \\
 & BaseNN & 0.860 & 0.447 & 0.807 & 0.541 & \textbf{0.840} & \textbf{0.726} & \underline{\textbf{1.000}} & \underline{\textbf{1.000}} \\
  \bottomrule
\end{tabular}
\caption{Results for counterfactual generation (mean over 5 seeds). The best value for each dataset is marked in bold.
ConTex outperforms the instance-based method ForecastCF in 16/16 cases and surpasses the retrieval-based reference BaseNN in 8/16 cases, demonstrating its ability to generate valid counterfactuals beyond existing dataset samples.
Due to target calibration (see Table \ref{full_coverage_comparison}), some trajectories already satisfy the constraints, allowing retrieval-based methods to achieve high validity. Accordingly, BaseNN performs equally (difference <1\%) in 3/16 cases and achieves the highest validity in 5/16 cases.}
\label{tab:valid_table}
\end{table}
 
 Table~\ref{tab:valid_table} reports the validity metrics of the generated counterfactuals. ConTex achieves high validity and S-AUC scores when producing counterfactuals across all datasets.
 
\begin{table}[!htbp]
\centering
\tiny
\setlength{\tabcolsep}{3pt}
\begin{tabular}{llcccccccc}
  \toprule
Model & CF Model & \multicolumn{2}{c}{NN5} & \multicolumn{2}{c}{Electricity} & \multicolumn{2}{c}{Tourism} & \multicolumn{2}{c}{M4} \\
  \midrule
 &  & Comp. $\uparrow$ & Prox. $\downarrow$ & Comp. $\uparrow$ & Prox. $\downarrow$ & Comp. $\uparrow$ & Prox. $\downarrow$ & Comp. $\uparrow$ & Prox. $\downarrow$ \\
  \midrule
\multirow{4}{*}{PatchTST} & ForecastCF & \textbf{0.455${\scriptstyle\pm0.186}$} & \underline{\textbf{2.613${\scriptstyle\pm1.853}$}} & 0.278${\scriptstyle\pm0.226}$ & \underline{\textbf{8.988${\scriptstyle\pm5.161}$}} & 0.295${\scriptstyle\pm0.198}$ & \textbf{8.012${\scriptstyle\pm3.372}$} & 0.862${\scriptstyle\pm0.051}$ & \underline{\textbf{0.299${\scriptstyle\pm0.139}$}} \\
 & ConTex & 0.314${\scriptstyle\pm0.111}$ & 13.129${\scriptstyle\pm0.552}$ & \textbf{0.450${\scriptstyle\pm0.024}$} & 21.510${\scriptstyle\pm0.690}$ & \textbf{0.582${\scriptstyle\pm0.059}$} & 12.233${\scriptstyle\pm0.488}$ & \textbf{0.974${\scriptstyle\pm0.001}$} & 1.239${\scriptstyle\pm0.006}$ \\
 & BaseShift & 0.031 & 13.591 & 0.024 & 41.263 & 0.016 & 35.101 & 0.034 & 4.811 \\
 & BaseNN & 0.034 & 17.099 & 0.078 & 50.833 & 0.117 & 28.390 & 0.789 & 0.507 \\
  \midrule
\multirow{4}{*}{N-HiTS} & ForecastCF & 0.366${\scriptstyle\pm0.130}$ & \textbf{4.130${\scriptstyle\pm2.509}$} & 0.275${\scriptstyle\pm0.234}$ & \textbf{14.195${\scriptstyle\pm8.889}$} & 0.309${\scriptstyle\pm0.204}$ & \textbf{7.540${\scriptstyle\pm2.455}$} & 0.687${\scriptstyle\pm0.109}$ & \textbf{0.367${\scriptstyle\pm0.133}$} \\
 & ConTex & \textbf{0.447${\scriptstyle\pm0.077}$} & 15.977${\scriptstyle\pm0.300}$ & \underline{\textbf{0.765${\scriptstyle\pm0.005}$}} & 20.315${\scriptstyle\pm0.390}$ & \textbf{0.517${\scriptstyle\pm0.055}$} & 11.067${\scriptstyle\pm0.640}$ & \textbf{0.977${\scriptstyle\pm0.000}$} & 0.747${\scriptstyle\pm0.016}$ \\
 & BaseShift & 0.031 & 12.961 & 0.018 & 55.874 & 0.017 & 34.987 & 0.033 & 4.760 \\
 & BaseNN & 0.041 & 14.606 & 0.086 & 53.817 & 0.128 & 28.669 & 0.595 & 0.831 \\
  \midrule
\multirow{4}{*}{DLinear} & ForecastCF & 0.460${\scriptstyle\pm0.132}$ & \textbf{5.195${\scriptstyle\pm3.870}$} & 0.263${\scriptstyle\pm0.222}$ & \textbf{13.010${\scriptstyle\pm8.741}$} & 0.425${\scriptstyle\pm0.181}$ & \textbf{10.580${\scriptstyle\pm5.471}$} & 0.765${\scriptstyle\pm0.083}$ & \textbf{0.612${\scriptstyle\pm0.275}$} \\
 & ConTex & \underline{\textbf{0.582${\scriptstyle\pm0.016}$}} & 19.164${\scriptstyle\pm0.509}$ & \textbf{0.607${\scriptstyle\pm0.008}$} & 18.754${\scriptstyle\pm0.243}$ & \underline{\textbf{0.704${\scriptstyle\pm0.014}$}} & 12.285${\scriptstyle\pm0.299}$ & \underline{\textbf{0.977${\scriptstyle\pm0.000}$}} & 0.751${\scriptstyle\pm0.013}$ \\
 & BaseShift & 0.031 & 12.967 & 0.018 & 54.826 & 0.017 & 34.977 & 0.034 & 4.757 \\
 & BaseNN & 0.030 & 17.424 & 0.075 & 52.012 & 0.109 & 28.325 & 0.586 & 0.836 \\
  \midrule
\multirow{4}{*}{TiDE} & ForecastCF & \textbf{0.268${\scriptstyle\pm0.111}$} & \textbf{3.481${\scriptstyle\pm1.707}$} & 0.258${\scriptstyle\pm0.227}$ & \textbf{12.427${\scriptstyle\pm8.370}$} & 0.283${\scriptstyle\pm0.199}$ & \underline{\textbf{6.230${\scriptstyle\pm1.792}$}} & 0.730${\scriptstyle\pm0.112}$ & \textbf{0.324${\scriptstyle\pm0.143}$} \\
 & ConTex & 0.249${\scriptstyle\pm0.040}$ & 10.179${\scriptstyle\pm0.148}$ & \textbf{0.690${\scriptstyle\pm0.003}$} & 18.072${\scriptstyle\pm0.092}$ & \textbf{0.411${\scriptstyle\pm0.026}$} & 8.342${\scriptstyle\pm0.726}$ & \textbf{0.948${\scriptstyle\pm0.020}$} & 1.719${\scriptstyle\pm0.991}$ \\
 & BaseShift & 0.031 & 13.187 & 0.019 & 57.002 & 0.016 & 35.142 & 0.033 & 4.782 \\
 & BaseNN & 0.046 & 13.200 & 0.080 & 52.895 & 0.118 & 27.989 & 0.570 & 0.823 \\
  \bottomrule
\end{tabular}
\caption{Results for counterfactual generation (mean over 5 seeds). Best reached value for every dataset is marked in bold. ConTex achieves the best compactness in 14 out of 16 cases.}
\label{tab:manifold_table}
\end{table}

We evaluate whether a method outperforms all baselines in both Validity Ratio and S-AUC across each of the 16 dataset–backbone combinations. Remarkably, instance-based generation (ForecastCF) is consistently outperformed by ConTex, and often even by the simple BaseShift heuristic, across all datasets and backbones. Notably, ConTex is the only method that frequently surpasses the retrieval-based reference BaseNN and achieves the strongest performance on average across datasets and forecasting backbones. Averaged over all dataset–backbone combinations, ConTex improves the mean Validity Ratio from 0.891 to 0.910 compared to BaseNN, corresponding to a relative improvement of approximately 2.1 \%. Furthermore, ConTex increases the mean S-AUC from 0.677 to 0.830, yielding an average improvement of approximately 22.6 \%. This indicates that it learns a global intervention policy that generalizes beyond dataset coverage, revealing how to reach trajectories even when they are not present in the historical data.

\begin{figure*}[!htbp]
    \centering

    \textbf{ConTex}

    \begin{tabular}{cc}
        \includegraphics[width=0.47\textwidth]{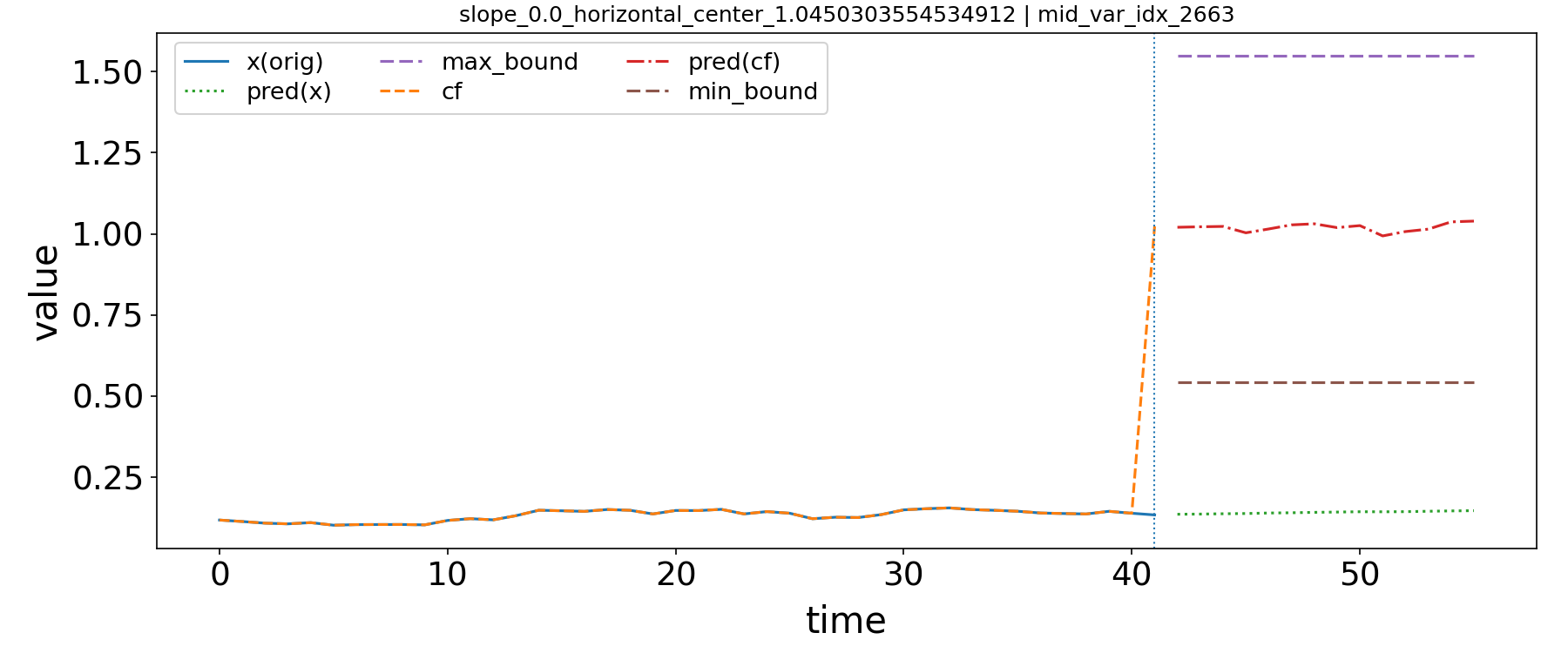} &
        \includegraphics[width=0.47\textwidth]{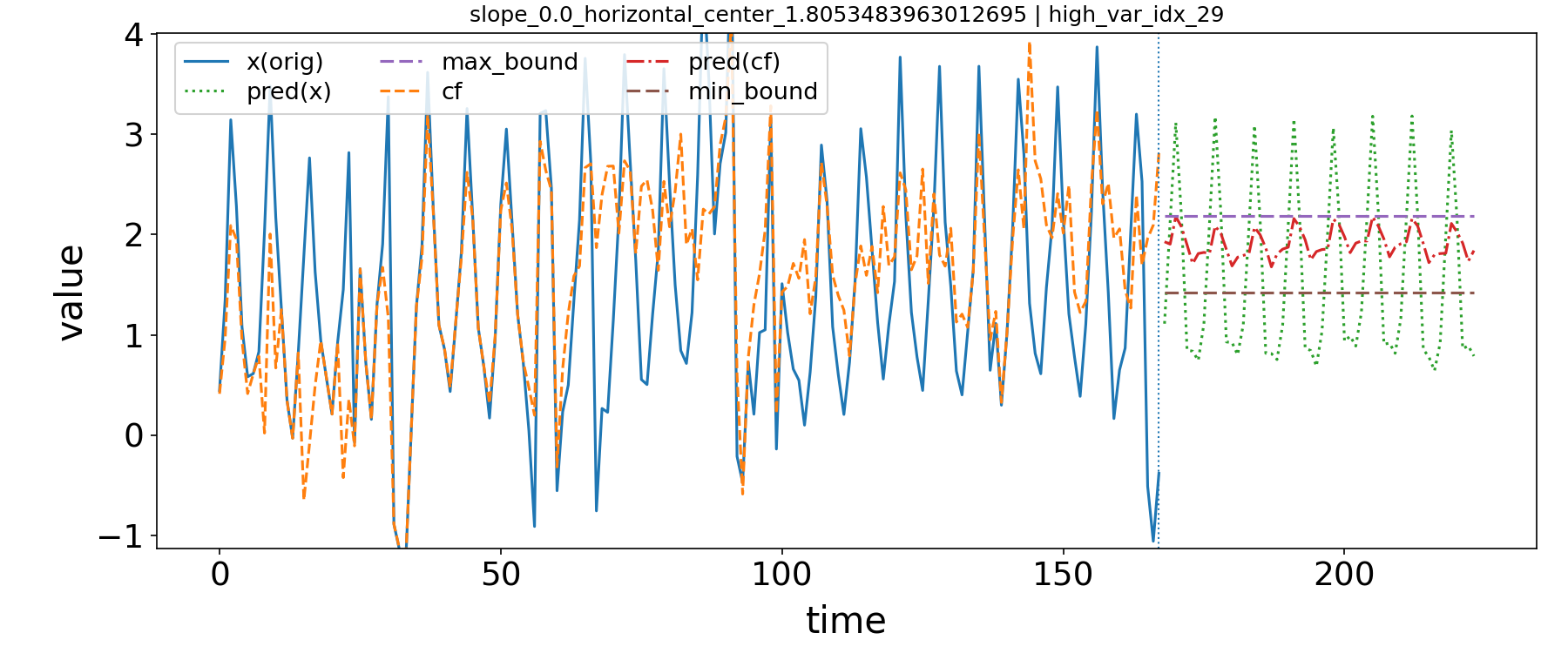}

    \end{tabular}

    \vspace{0.5em}

    \textbf{ForecastCF}

    \begin{tabular}{cc}
        \includegraphics[width=0.47\textwidth]{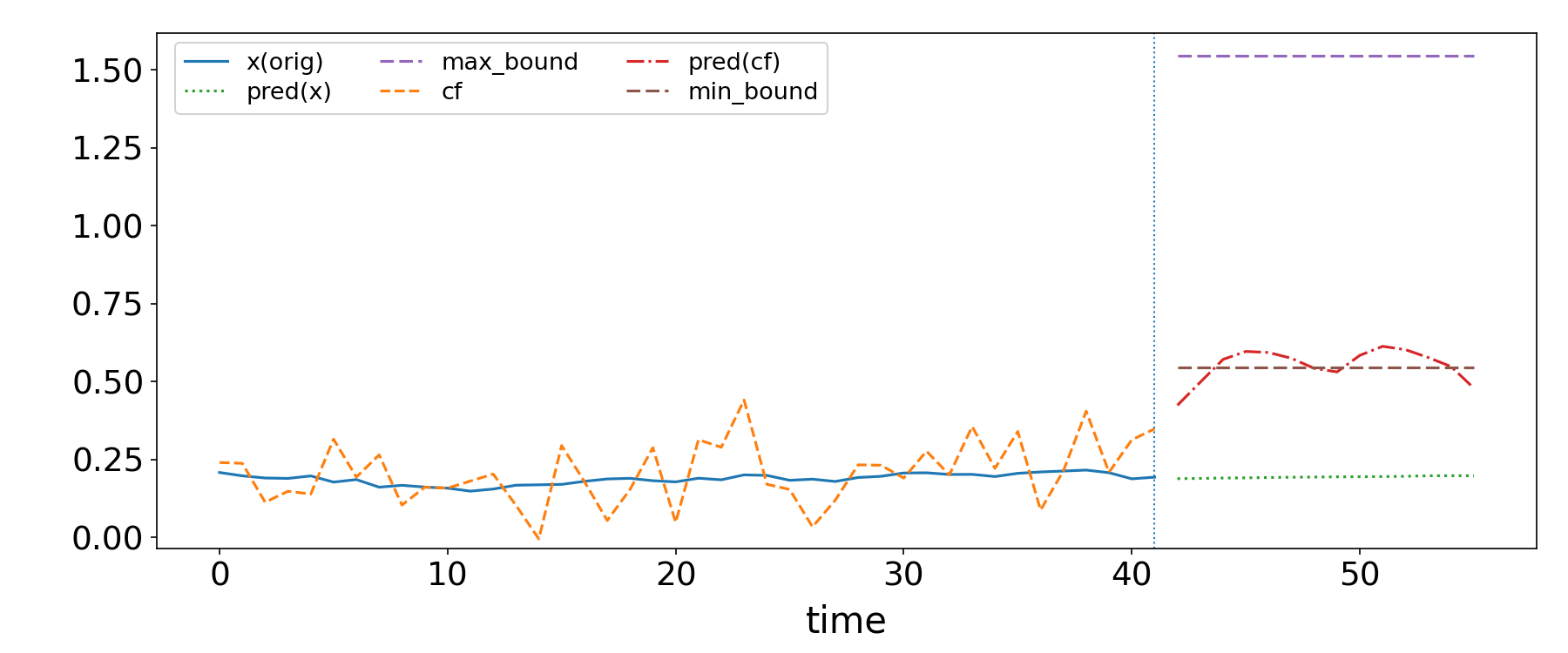} &
        \includegraphics[width=0.47\textwidth]{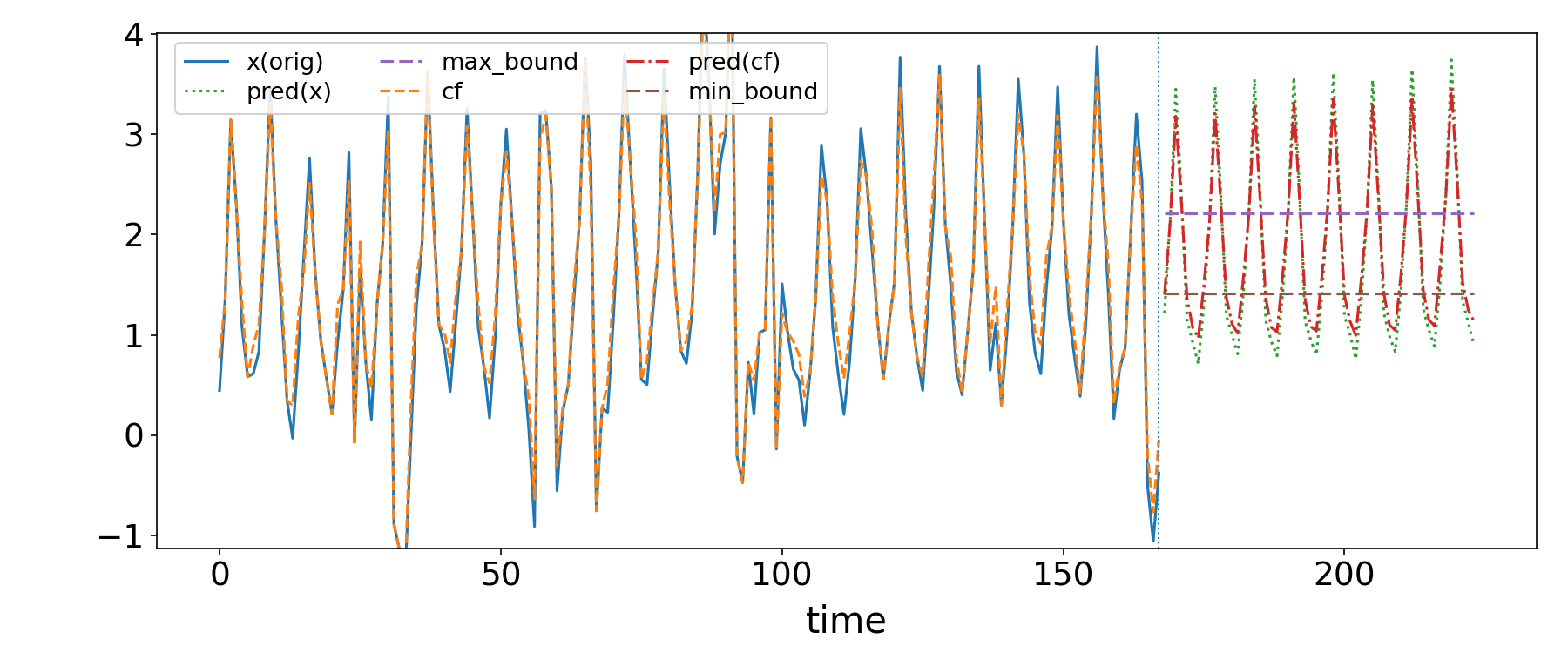}
    \end{tabular}
     \caption{
    Qualitative comparison of generated counterfactuals on M4 and NN5 using N-HiTS. Across both low- and high-noise datasets, the instance-based method ForecastCF produces numerous distributed changes. In contrast, ConTex applies interventions only when necessary: on smooth data, it focuses on recent timesteps, while on noisier data it introduces more distributed adjustments to achieve the desired trajectory. ConTex further shows higher precision in aligning forecasts with the target trend.
    }
    \label{fig:qualitative_samples}
\end{figure*}

Beyond validity, we further analyze the sparsity of the generated interventions in Table~\ref{tab:manifold_table}. ConTex achieves higher compactness consistently, requiring fewer modified timesteps. It exhibits predominantly lower proximity than BaseShift and BaseNN, though not quite as low as instance-based generation. However, this trade-off should be interpreted in context, as proximity is only meaningful among valid counterfactuals. Methods that produce small but ineffective perturbations can lead to misleadingly strong results in terms of proximity and compactness. Furthermore, for ConTex, the trade-off between proximity and validity is a controllable design choice. By increasing the weight of the proximity term and relaxing sparsity constraints, ConTex can generate counterfactuals with substantially smaller deviations at the cost of reduced validity or compactness. This occurs because stronger proximity regularization constrains deviations from the original trajectory, limiting the interventions needed for validity (e.g., see \ref{appendix:proximity:tradeoff}).

\begin{figure*}[!htbp]
    \centering
\begin{tabular}{cc}
        \includegraphics[width=0.47\textwidth]{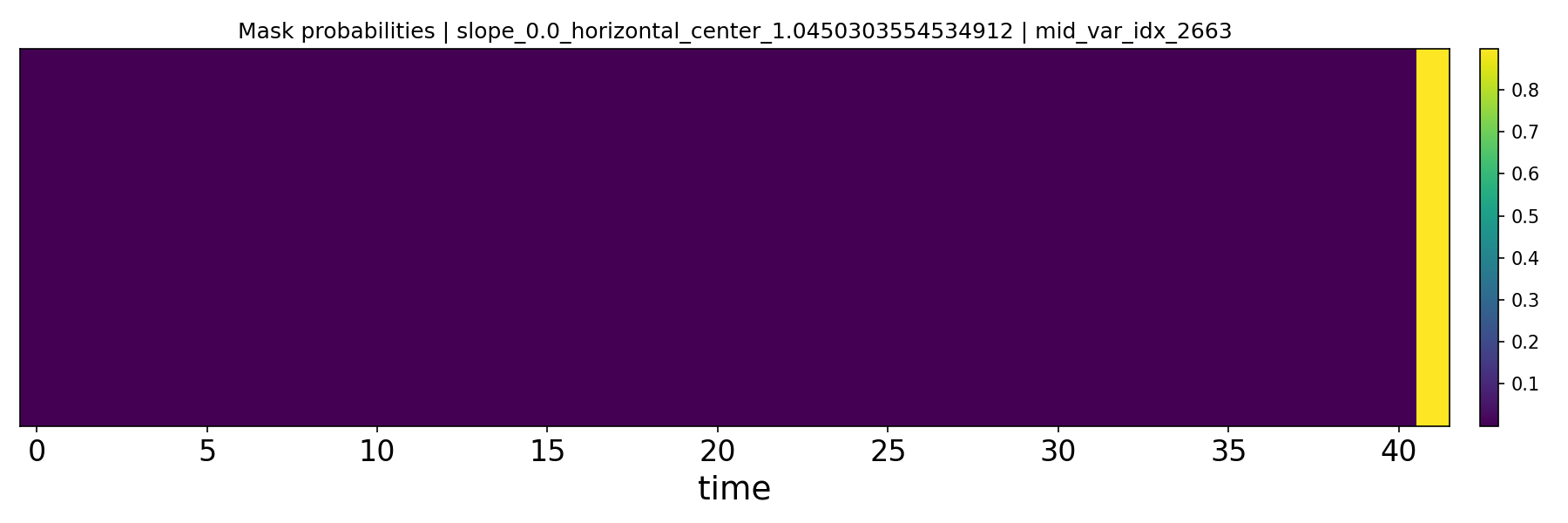} &
        \includegraphics[width=0.47\textwidth]{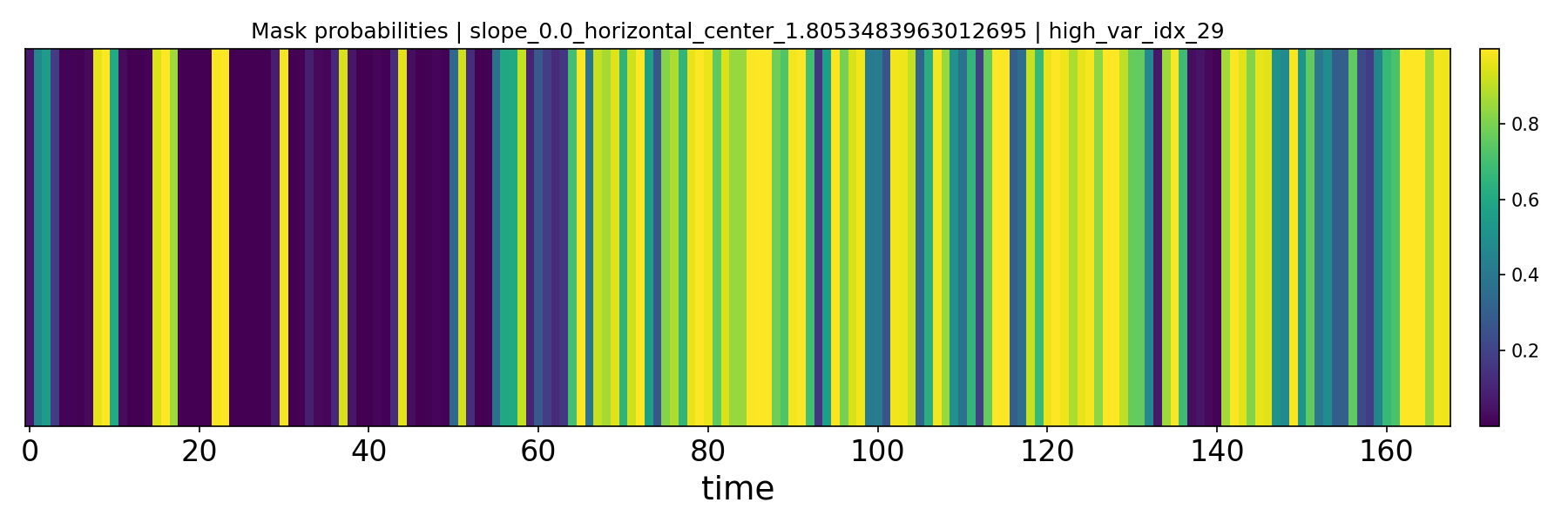} \\
\end{tabular}
\caption{Interventional (target-conditioned) attribution: Temporal relevance masks for a smooth M4 sample (left) compared to a more dispersed intervention pattern on NN5 (right). While for both datasets the intervention probability increases for more recent inputs, ConTex is able to learn that smooth data requires only minimal changes while noisy data requires substantially broader temporal structures, resulting in more distributed interventions across the input sequence.}
    \label{fig:mask}
\end{figure*}

Another important aspect is the behavior of the methods across datasets with varying characteristics. ConTex adapts to the underlying data properties, as illustrated in Figure~\ref{fig:qualitative_samples}. This is further reflected in the interventional relevance mask, which represents the  importance of each timestep as a probability between 0 and 1 to reach the desired target. For the horizontal target trends considered above, we visualize the resulting masks in Figure~\ref{fig:mask}.

\begin{figure*}[!htbp]
    \centering
    \setlength{\tabcolsep}{2pt}
    \renewcommand{\arraystretch}{1.1}
    \small
    \begin{tabular}{@{}lcccc@{}}
        & \textbf{Ratio} $\uparrow$
        & \textbf{S-AUC} $\uparrow$
        & \textbf{Compactness} $\uparrow$
        & \textbf{Proximity} $\downarrow$ \\[-0.3em]

        \rotatebox[origin=c]{90}{\textbf{ConTex}}
        & \includegraphics[width=0.225\textwidth, trim=0 5 0 25, clip]{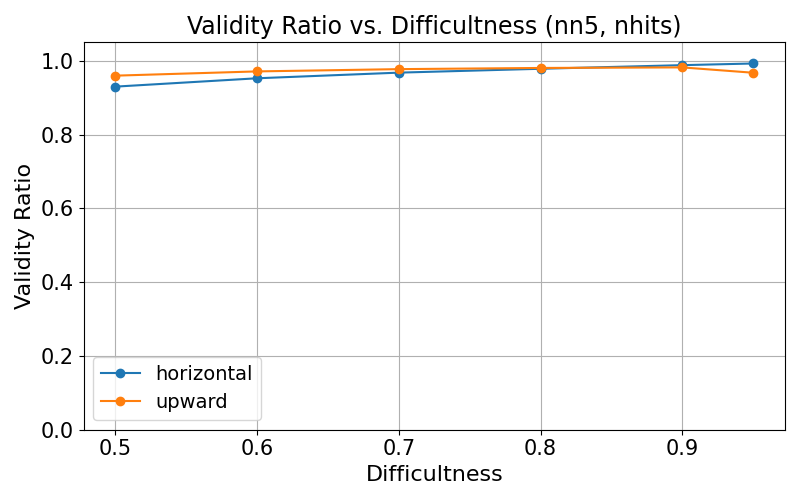}
        & \includegraphics[width=0.225\textwidth, trim=0 5 0 25, clip]{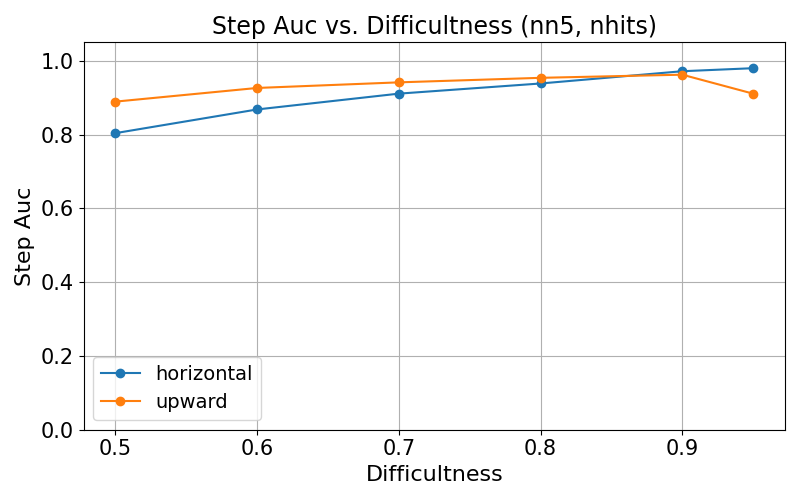}
        & \includegraphics[width=0.225\textwidth, trim=0 5 0 25, clip]{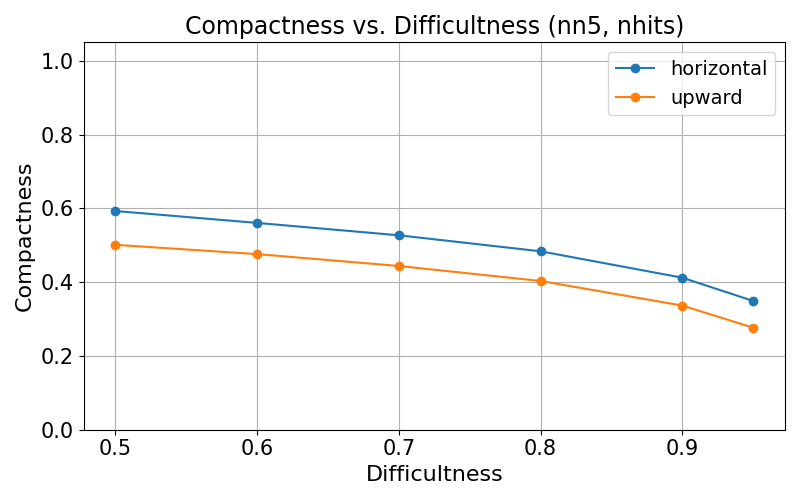}
        & \includegraphics[width=0.225\textwidth, trim=0 5 0 25, clip]{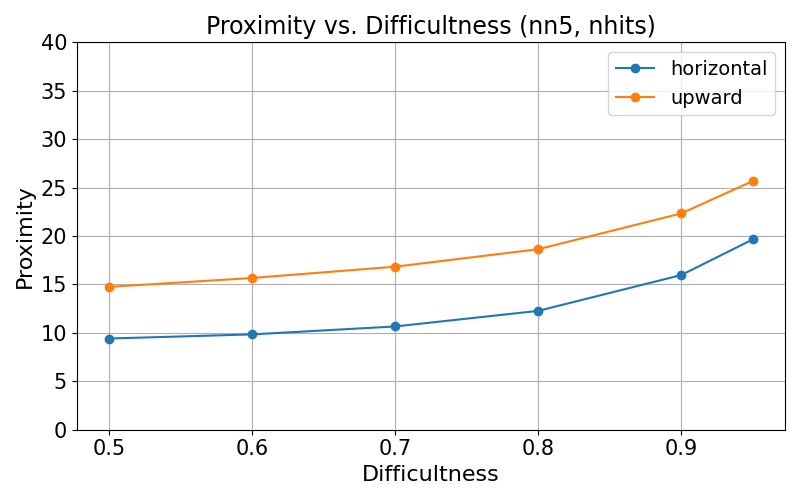} \\[-0.2em]

        \rotatebox[origin=c]{90}{\textbf{ForecastCF}}
        & \includegraphics[width=0.225\textwidth, trim=0 5 0 25, clip]{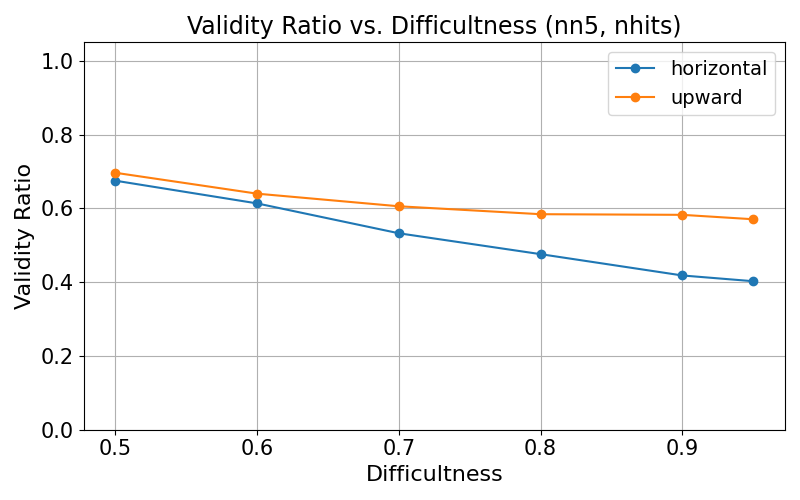}
        & \includegraphics[width=0.225\textwidth, trim=0 5 0 25, clip]{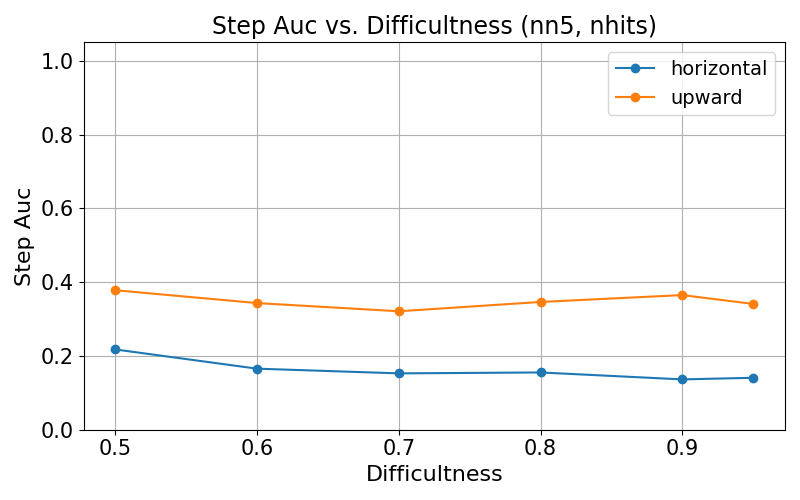}
        & \includegraphics[width=0.225\textwidth, trim=0 5 0 25, clip]{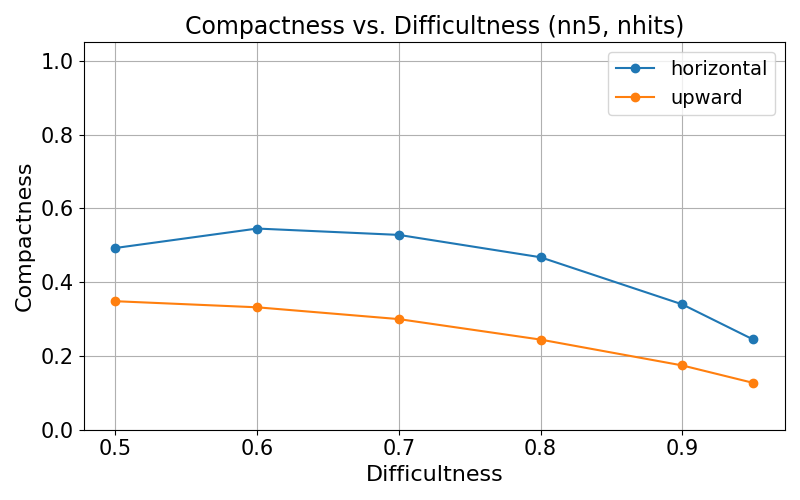}
        & \includegraphics[width=0.225\textwidth, trim=0 5 0 25, clip]{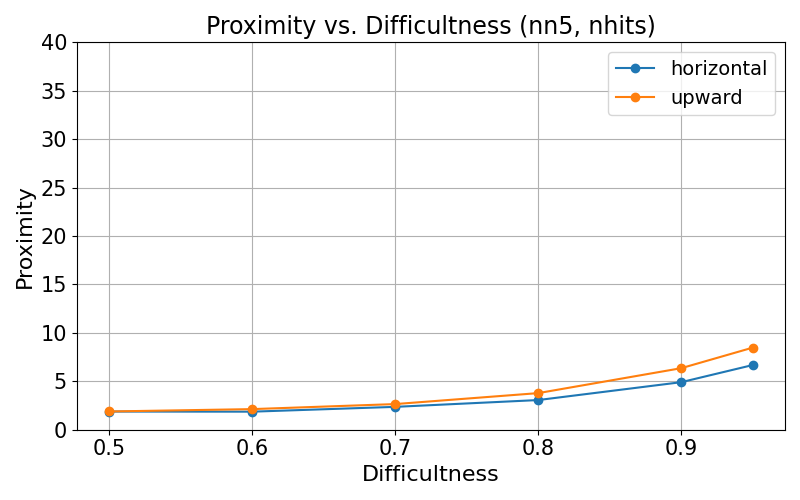}
    \end{tabular}
    \caption{Metric behavior across increasing target difficulty on NN5 with N-HiTS.}
    \label{fig:metric_behavior_nn5_nhits}
\end{figure*}

Despite methodological differences, both ConTex and ForecastCF exhibit similar scaling behavior with increasing target difficulty (Figure~\ref{fig:metric_behavior_nn5_nhits}). Harder targets generally require larger and less compact interventions, reflected by decreasing compactness and increasing proximity. However, while ConTex maintains stable validity across difficulty regimes, ForecastCF shows a substantial degradation in validity, particularly for horizontal targets.

\begin{figure*}[!htbp]
    \centering
\begin{tabular}{cc}
        \includegraphics[width=0.47\textwidth]{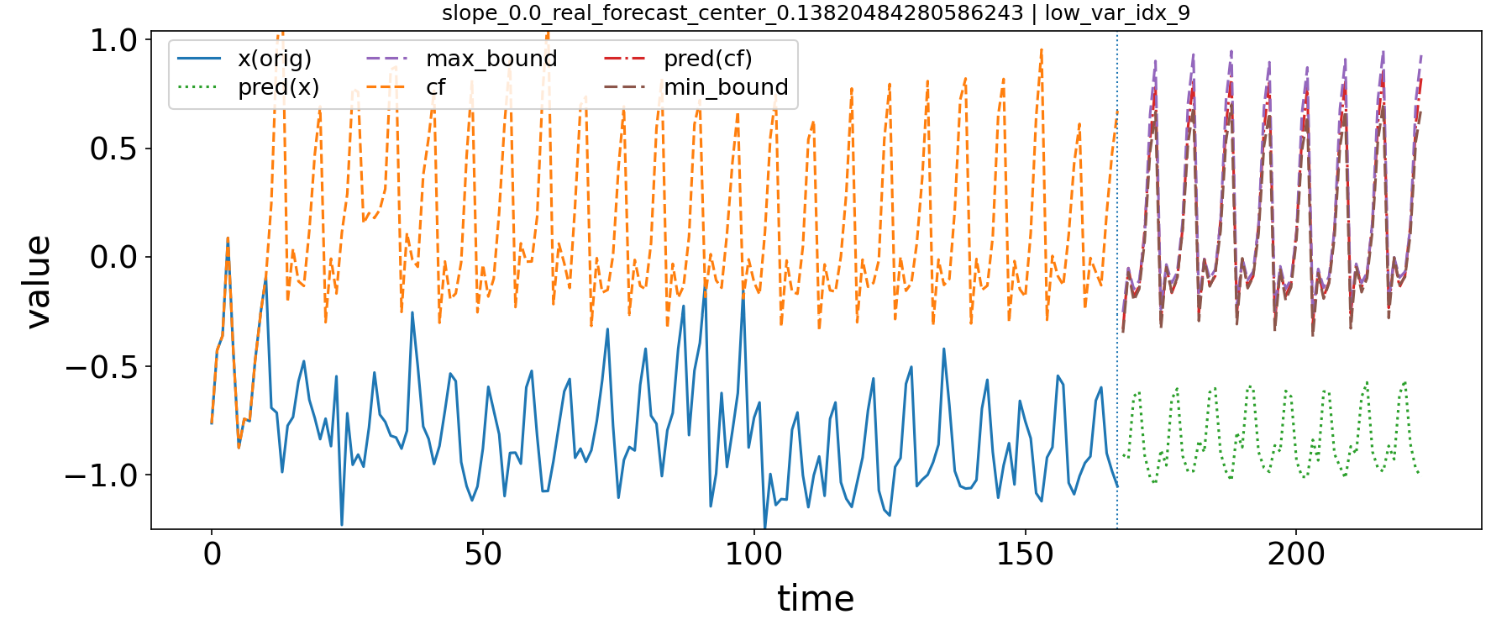} &
        \includegraphics[width=0.47\textwidth]{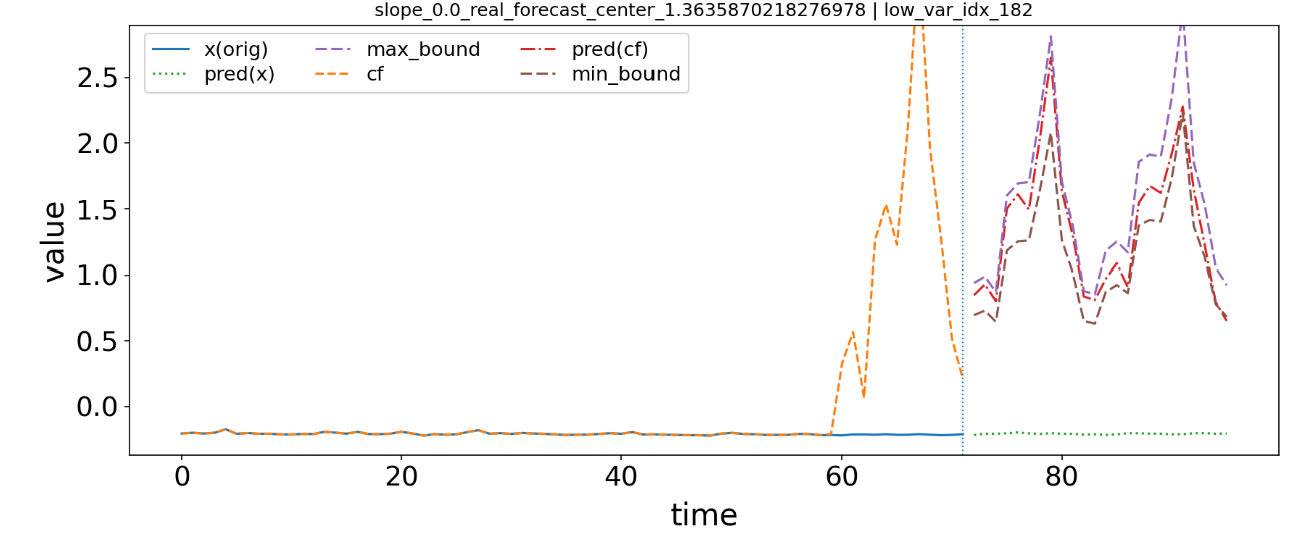} \\
\end{tabular}
\caption{Qualitative examples on realistic target trajectories. ConTex remains capable of reproducing complex target patterns beyond standardized trend-shift benchmarks.}
    \label{real_patterns}
\end{figure*}

\paragraph{Generalization to Realistic Target Patterns} Beyond standardized target families, we additionally evaluate ConTex on realistic target patterns sampled from the data distribution. Two examples are given in Figure \ref{real_patterns}. Both represent challenging distributional shifts involving substantial differences in temporal structure, amplitude, and overall signal dynamics. Despite these shifts, ConTex remains capable of generating highly accurate and valid counterfactual trajectories.

\paragraph{Runtime Study.} 
All experiments were conducted on an NVIDIA Tesla V100 GPU (32 GB VRAM) with CUDA 12.4 (driver 550.163.01). We report a representative study on NN5 using N-HiTS in the Appendix \ref{appendix:runtime}. ConTex achieves a near-constant inference time of $\sim$0.007 s per sample. Even in the most challenging setting (95th percentile), this corresponds to a speedup of $3.7$–$4.2$ orders of magnitude compared to instance-based generation (108 s per sample), whose runtime increases with target difficulty. In the least challenging setting, ConTex remains several orders faster (e.g., 32.5 s vs. 0.007 s), demonstrating consistent gains across all difficulty levels. Accounting for training (Table \ref{tab:contexruntime}), ConTex incurs a one-time cost of 5–70 minutes depending on the dataset and the configuration (10–20 minutes on M4, up to 60–70 minutes on NN5). In contrast, ForecastCF requires $\sim$18 hours for all NN5 settings across 5 seeds (3.75 days total vs. 5.8 hours), resulting in an overall reduction in computational cost of at least $12\times$–$36\times$.

\subsection{Ablation Study}

\begin{table}[!htbp]
  \centering
  \begin{tabular}{lcccc}
  \toprule
Variant & Ratio $\uparrow$ & S-AUC $\uparrow$ & Comp. $\uparrow$ & Prox. $\downarrow$ \\
  \midrule
ConTex Full & 0.971 $\pm$ 0.026 & 0.921 $\pm$ 0.062 & 0.447 $\pm$ 0.077 & 15.977 $\pm$ 0.300 \\
w/o $\mathcal{L}_{L1}$ (MSE) & 0.930 $\pm$ 0.036 & 0.900 $\pm$ 0.049 & 0.100 $\pm$ 0.123 & 12.953 $\pm$ 1.268 \\
w/o Cond. Encoder & 0.346 $\pm$ 0.003 & 0.136 $\pm$ 0.008 & 0.679 $\pm$ 0.045 & 7.691 $\pm$ 0.606 \\
w/o Decomposition & 0.998 $\pm$ 0.001 & 0.990 $\pm$ 0.002 & 0.340 $\pm$ 0.043 & 15.497 $\pm$ 0.273 \\
w/o $\hat{y}$ Condition  & 0.999 $\pm$ 0.001 & 0.995 $\pm$ 0.002 & 0.201 $\pm$ 0.061 & 15.175 $\pm$ 0.153 \\
  \bottomrule
  \end{tabular}
  \caption{Ablation study on NN5 with N-HiTS. Values are reported as mean $\pm$ standard deviation across seeds.}
  \label{tab:ablation_nn5_nhits}
\end{table}

We evaluated each architectural design choice through four ablation studies, each averaged over five random seeds. Since the architecture was optimized on NN5 using N-HiTS, we use this configuration as a representative example. The corresponding ablation results on Tourism (Table~\ref{tab:ablation_tourism_nhits}) confirm the same trends.

First, we prove our decision to use MAE for optimizing compactness by replacing it with MSE for $\mathcal{L}{\mathrm{prox}}$. Due to stronger penalties on large deviations, MSE discourages localized changes, reducing compactness (w/o $\mathcal{L}{L1}$). Accordingly, we observe substantial drops in compactness, validity ratio, and S-AUC. As a second ablation, we remove the conditioning encoder. This substantially impairs the model’s ability to differentiate between desired target outcomes, leading to a substantial decrease in both validity ratio and S-AUC. At the same time, this result highlights that proximity and compactness must be interpreted with caution: although these metrics may appear to improve, such improvements are not meaningful if the method fails to achieve sufficiently high validity.

As a third ablation, we remove the decomposition into mask and strength. While this encourages ConTex to explore modifications that lead to improvements in validity of around three percent more extensively, it drastically reduces compactness. Since sparse and interpretable modifications are of central relevance to counterfactuals, this trade-off is not preferable. As a final ablation, we remove $\hat{y}$ from the condition input. Interestingly, this does not adversely affect validity, but results in a substantial reduction in compactness. This indicates that incorporating the current forecast is important for guiding the model toward sparser intervention strategies.

\section{Conclusion}
\label{Limitations}
We demonstrate that, for time series forecasting, instance-wise optimization for counterfactual generation can be effectively replaced by amortized inference based on a global intervention policy. Furthermore, a structurally decomposed intervention function, separating where to intervene from how, yields counterfactuals that are more valid, sparser, and orders of magnitude cheaper to generate, outperforming instance-based optimization and often even realistic data samples. Additionally, the decomposed architecture of ConTex provides insights into target-conditioned feature relevance and model behavior, offering a forward-looking, intervention-based perspective that is not captured by standard post-hoc attribution methods. 

While ConTex demonstrates strong performance, limitations remain. Firstly, ConTex does not explicitly enforce plausibility constraints, but instead implicitly promotes temporal coherence through regularization induced by the training data distribution, temporal encoding, and minimal-intervention objectives. Secondly, although the architecture generalizes across datasets and forecasting models, some hyperparameters, particularly those controlling proximity and sparsity, still require limited dataset-specific tuning, varying within relatively narrow ranges (0.05–0.4 for sparsity and 0.4–0.5 for proximity). Addressing these limitations, particularly through realistic constraints and more complex target formulations, represents an important direction for future work.

\section{Acknowledgments}{
This work was supported by the European Union through the European Regional Development Fund (ERDF/EFRE) under grant number EFRE-20800908. The sole responsibility for the content of this publication lies with the authors.}


\newpage

\appendix

\section{Technical appendices and supplementary material}

\subsection{Architecture Setup}
\label{appendix:hyperparameter}

\subsubsection{General Configuration}
\label{full_implementation}

\begin{table}[!htbp]
\centering
\begin{tabular}{l l}
\toprule
\textbf{Component} & \textbf{Configuration} \\
\multicolumn{2}{l}{\textit{Temporal Encoder}} \\
Encoder Type & TCN \\
Encoding Dimension & 64 \\
Kernel Size & 3 \\
TCN Dropout & 0.1 \\
Receptive Field Ratio & 1.0 \\
Dilations & Exponentially increasing, $2^i$ \\

\midrule
\multicolumn{2}{l}{\textit{Condition Encoder}} \\
Condition Input & Target parameters + forecast status \\
Conditioning MLP & 2 layers, hidden dim 8, embed dim 16 \\
Condition Injection & FiLM + temporal concatenation \\
FiLM Scaling & bounded, scale = 0.5 \\

\midrule
\multicolumn{2}{l}{\textit{Prediction Heads}} \\
Mask Head & Time-distributed MLP: 32--16--1, sigmoid \\
Strength Head & Time-distributed MLP: 32--16--$D$, tanh \\
Strength Scaling & Conditioned StrengthScaler, initial scale 1.0 \\
Intervention & $z = m \odot s$ \\
Output & $x^{\mathrm{cf}} = x + z$ \\
\midrule
\multicolumn{2}{l}{\textit{Training}} \\
Batch Size & 128 \\
Optimizer & Adam \\
Seeds & 0–4 \\
\midrule
\multicolumn{2}{l}{\textit{Loss Weights}} \\
$\lambda_{\text{center}}$ & 0.2 \\
$\lambda_{\text{valid}}$ & 1.0 \\
\midrule
\multicolumn{2}{l}{\textit{Sparsity}} \\
Threshold $\tau$ & 0.05 \\
Annealing & 0.1 $\rightarrow$ 1.0 (10 epochs) \\
\bottomrule
\end{tabular}
\caption{Architectural configuration of ConTex. To ensure stable convergence, small deviations in the counterfactuals below a threshold of $\tau = 0.05$ are ignored. Furthermore, we employ sparsity annealing, gradually increasing the sparsity regularization during training to encourage the exploration of diverse intervention patterns in early epochs while promoting more compact solutions later on.}
\end{table}

\newpage
\subsubsection{Dataset Specific}

\subsubsubsection{}
\begin{table}[!htbp]
\centering
\begin{tabular}{lccccc}
\toprule
\textbf{Dataset} & \textbf{Learning Rate} & \textbf{TCN Dropout} & \textbf{Cond. Dropout} & $\lambda_{\text{sparse}}$ & $\lambda_{\text{prox}}$ \\
\midrule
Electricity & 2e-4 & 0.10 & 0.00 & 0.05 & 0.50 \\
NN5         & 1e-4 & 0.05 & 0.00 & 0.02 & 0.40 \\
M4          & 1e-3 & 0.05 & 0.05 & 0.40 & 0.40  \\
Tourism     & 1e-4 & 0.10 & 0.00 & 0.05 & 0.50 \\
\bottomrule
\end{tabular}
\caption{Dataset-specific hyperparameters for ConTex. A small set of optimization parameters is tuned for each dataset, while all other architectural components remain fixed.}
\end{table}

\begin{table}[!htbp]
\centering
\small
\begin{tabular}{l l l l}
\toprule
\textbf{Dataset} & \textbf{Train} & \textbf{Validation} & \textbf{Test} \\
\midrule
NN5 & 85\% of training series & 10\% of training series & Last 56 days of each series \\
M4 & 85\% of training series & 10\% of training series & Last 14 days of each series \\
Tourism & 85\% of training series & 10\% of training series & Last 24 months of each series \\
Electricity & 85\% of training series & 10\% of training series & Last 4 months of each series \\
\bottomrule
\end{tabular}
\caption{Dataset-specific data splits. For all datasets, training and validation are constructed from the provided training portion of each dataset, while test samples correspond to the final forecasting horizon of each series.}
\label{tab:data_splits}
\end{table}

\newpage
\subsection{Target Calibration}

\subsubsection{Calibration Example}
\label{dlinearcalibration}

\begin{table}[!htbp]
  \centering
  \begin{tabular}{lcccc}
    \toprule
    Target & NN5 & Electricity & Tourism & M4 \\
    \midrule
    0.5 horizontal & 0.479 & 0.508 & 0.526 & 0.500 \\
    0.5 upward & 0.268 & 0.220 & 0.145 & 0.489 \\
    0.6 horizontal & 0.403 & 0.346 & 0.236 & 0.475 \\
    0.6 upward & 0.212 & 0.152 & 0.092 & 0.475 \\
    0.7 horizontal & 0.306 & 0.166 & 0.092 & 0.466 \\
    0.7 upward & 0.159 & 0.089 & 0.054 & 0.464 \\
    0.8 horizontal & 0.202 & 0.066 & 0.035 & 0.413 \\
    0.8 upward & 0.107 & 0.046 & 0.027 & 0.408 \\
    0.9 horizontal & 0.103 & 0.017 & 0.010 & 0.321 \\
    0.9 upward & 0.049 & 0.014 & 0.009 & 0.315 \\
    0.95 horizontal & 0.048 & 0.005 & 0.003 & 0.226 \\
    0.95 upward & 0.020 & 0.005 & 0.002 & 0.220 \\
    \bottomrule
  \end{tabular}
  \caption{Example for calibration of target bounds across datasets on DLinear. The bounds width is adjusted so that forecasting purely based on existing dataset samples leads to approximately a 50 \%  validity ratio on horizontal bounds in the common data regime, ensuring feasibility on each dataset and ensuring that, at least in common regimes, samples already exist that satisfy the constraints to a certain degree.}
\end{table}

\subsubsection{Full Target Coverage Comparison}
\label{full_coverage_comparison}

\begin{table}[ht]
  \centering
  \begin{tabular}{lcccc}
    \toprule
    & \multicolumn{2}{c}{Electricity} & \multicolumn{2}{c}{M4} \\
    \cmidrule(lr){2-3} \cmidrule(lr){4-5}
    Target & Ratio & Full & Ratio & Full \\
    \midrule
    0.5 horizontal & 0.500 & 0.292 & 0.500 & 0.493 \\
    0.5 upward & 0.150 & 0.000 & 0.492 & 0.475 \\
    0.6 horizontal & 0.354 & 0.099 & 0.477 & 0.470 \\
    0.6 upward & 0.110 & 0.000 & 0.477 & 0.469 \\
    0.7 horizontal & 0.178 & 0.012 & 0.468 & 0.461 \\
    0.7 upward & 0.071 & 0.000 & 0.466 & 0.458 \\
    0.8 horizontal & 0.067 & 0.000 & 0.412 & 0.405 \\
    0.8 upward & 0.039 & 0.000 & 0.409 & 0.400 \\
    0.9 horizontal & 0.019 & 0.000 & 0.318 & 0.312 \\
    0.9 upward & 0.013 & 0.000 & 0.314 & 0.307 \\
    0.95 horizontal & 0.005 & 0.000 & 0.221 & 0.215 \\
    0.95 upward & 0.005 & 0.000 & 0.217 & 0.209 \\
    \bottomrule
  \end{tabular}
  \caption{Comparison for full target coverage for M4 and Electricity. While both were calibrated to a 50 percent validity ratio, the dataset distribution significantly differs, leading to a high percentage of samples that satisfy the target constraints on M4 with 100 \% validity, and therefore ensuring a high mean validity of BaseNN, while only a few satisfy the corresponding target constraints on Electricity.}
\end{table}

\newpage
\subsection{Evaluation Metrics}
\label{EvaluationMetrics}

We evaluate counterfactual quality using validity and data manifold closeness, adapted from prior work on time series counterfactuals \cite{wang_counterfactual_2023}.

\paragraph{Validity.}
The \emph{Validity Ratio} measures the fraction of forecasted timesteps satisfying the target constraints:
\begin{equation}
\text{ValidityRatio} =
\frac{1}{K} \sum_{k=1}^{K}
\left( \frac{1}{T} \sum_{t=1}^{T}
\mathbb{1} \left[ \hat{y}^{\,\prime}_{k,t} \in \mathcal{Y}_{\text{target},t} \right]
\right),
\end{equation}
where $K$ is the number of counterfactuals and $T$ the forecasting horizon.
We further report \emph{Stepwise Validity AUC}, capturing consecutive valid forecast steps.

\paragraph{Data manifold closeness.}
To assess how much counterfactuals deviate from the original input, we use \emph{Proximity} to measure the average Euclidean distance between the original input $x$ and the counterfactual $x'$, and \emph{Compactness} to measure the proportion of timesteps that remain unchanged:
\begin{equation}
\text{Proximity} =
\frac{1}{K} \sum_{k=1}^{K} \| x_k - x'_k \|,
\quad
\text{Compactness} =
\frac{1}{K} \sum_{k=1}^{K}
\left( \frac{1}{T} \sum_{t=1}^{T}
\mathbb{1} \left[ |x_{k,t} - x'_{k,t}| \leq \tau \right]
\right).
\end{equation}

\subsection{BaseNN}
\label{appendix:basenn}

\begin{algorithm}[!htbp]
\caption{BaseNN (adapted): Retrieval of samples satisfying the target trajectory. Candidate samples are selected based on constraint satisfaction, and the nearest neighbor is returned as the counterfactual.}
\begin{algorithmic}[1]
\Require Training data $x_{\mathrm{train}}$, model $f_\theta$, bounds $[L,U]$, query $x$, minimum candidates $M$
\State Compute predictions $\hat{y}_i = f_\theta(x_i)$ for all $x_i \in x_{\mathrm{train}}$
\State Compute validity scores:
\[
v_i = \sum_{t=1}^{H} \mathbb{1}\!\left[\hat{y}_{i,t} \in [L_t,U_t]\right]
\]
\For{$\tau \in \{H, H-1, \dots, 0\}$}
    \State $\mathcal{C}_\tau = \{x_i \mid v_i \geq \tau\}$
    \If{$|\mathcal{C}_\tau| \geq M$}
        \State \textbf{break}
    \EndIf
\EndFor
\State \Return $x^{\mathrm{cf}} = \arg\min_{x_i \in \mathcal{C}_\tau} \|x - x_i\|_2$
\end{algorithmic}
\end{algorithm}

\newpage
\subsection{Runtime Results}
\label{appendix:runtime}

\subsubsection{NN5: Generation Time}

\begin{table}[!htbp]
\centering
\small
\begin{tabular}{cll|rrrr}
\toprule
Difficulty & Direction & Method & Mean & Min & Max & Total (s) \\
\midrule
\multirow{4}{*}{0.50} & \multirow{2}{*}{horizontal} & ConTex & 0.007188 & 0.006870 & 0.007669 & 0.6627 \\
 &  & ForecastCF & 32.500600 & 7.210600 & 204.944300 & 3055.0566 \\
 & \multirow{2}{*}{upward} & ConTex & 0.007094 & 0.006814 & 0.008191 & 0.6646 \\
 &  & ForecastCF & 23.148600 & 7.205000 & 114.638600 & 2175.9652 \\
\midrule
\multirow{4}{*}{0.60} & \multirow{2}{*}{horizontal} & ConTex & 0.007076 & 0.006819 & 0.007998 & 0.6633 \\
 &  & ForecastCF & 15.892400 & 2.542100 & 209.665600 & 1493.8873 \\
 & \multirow{2}{*}{upward} & ConTex & 0.007057 & 0.006855 & 0.007852 & 0.6618 \\
 &  & ForecastCF & 26.798000 & 7.067000 & 80.611900 & 2519.0121 \\
\midrule
\multirow{4}{*}{0.70} & \multirow{2}{*}{horizontal} & ConTex & 0.007076 & 0.006807 & 0.007839 & 0.6636 \\
 &  & ForecastCF & 44.604500 & 6.970400 & 157.413000 & 4192.8264 \\
 & \multirow{2}{*}{upward} & ConTex & 0.007049 & 0.006837 & 0.007913 & 0.6609 \\
 &  & ForecastCF & 30.648600 & 6.848100 & 98.503000 & 2880.9678 \\
\midrule
\multirow{4}{*}{0.80} & \multirow{2}{*}{horizontal} & ConTex & 0.007040 & 0.006831 & 0.007805 & 0.6603 \\
 &  & ForecastCF & 60.288000 & 6.903700 & 186.274000 & 5667.0713 \\
 & \multirow{2}{*}{upward} & ConTex & 0.007074 & 0.006832 & 0.007850 & 0.6634 \\
 &  & ForecastCF & 50.014400 & 7.096800 & 121.303200 & 4701.3529 \\
\midrule
\multirow{4}{*}{0.90} & \multirow{2}{*}{horizontal} & ConTex & 0.007056 & 0.006819 & 0.007823 & 0.6617 \\
 &  & ForecastCF & 81.564700 & 11.216600 & 180.628700 & 7667.0773 \\
 & \multirow{2}{*}{upward} & ConTex & 0.007078 & 0.006878 & 0.007855 & 0.6638 \\
 &  & ForecastCF & 79.499700 & 8.407300 & 164.462500 & 7472.9678 \\
\midrule
\multirow{4}{*}{0.95} & \multirow{2}{*}{horizontal} & ConTex & 0.007065 & 0.006856 & 0.007832 & 0.6626 \\
 &  & ForecastCF & 108.473000 & 6.987100 & 266.547600 & 10196.4579 \\
 & \multirow{2}{*}{upward} & ConTex & 0.007065 & 0.006857 & 0.007905 & 0.6624 \\
 &  & ForecastCF & 104.345200 & 7.064400 & 223.090500 & 9808.4527 \\
\bottomrule
\end{tabular}
\caption{Runtime statistics by difficulty percentile, direction, and method. 
For ConTex, inference time is measured as the forward-pass runtime, 
including synchronization via NumPy conversion to ensure that all GPU operations are completed.}
\label{tab:runtimeappendix}
\end{table}

\subsubsection{ConTex Training Runtime Comparison}
\label{contextraining}

\begin{table}[h]
\centering
\begin{tabular}{lcc}
\toprule
\textbf{Component} & \textbf{NN5 (s)} & \textbf{M4 (s)} \\
\midrule
Forecast Model Preparation & 3.23 & 22.01 \\
Bounds Calibration        & 23.41 & 57.48 \\
\midrule
ConTex Training   & \textbf{3074.57} & \textbf{314.58} \\
Validation                & 1248.20 & 106.02 \\
\midrule
Test Evaluation           & 3.42 & 67.19 \\
Plot Generation           & 129.29 & 130.08 \\
\midrule
\textbf{Total Pipeline}   & 4542.80 & 758.34 \\
\bottomrule
\end{tabular}
\caption{Runtime breakdown of the full ConTex pipeline with N-HiTS on NN5 in comparison to M4. While the total runtime includes all stages in the pipeline such as model preparation, bounds calibration, evaluation, and visualization, we focus on the core training time (highlighted in bold) for comparison, excluding validation time between epochs. While training time on ConTex in general is quite low, e.g. on M4 around 5-10 minutes, it increases with more complex and noisy data patterns such as on NN5 requiring approximately 58 minutes for the reported runtime study.}
\label{tab:contexruntime}
\end{table}

\newpage
\subsection{Forecasting Models}
\label{appendix:smape}
\begin{table}[!htbp]
  \centering
  \footnotesize
  \begin{tabular}{lcccc}
    \toprule
    Model & NN5 & Electricity & Tourism & M4 \\
    \midrule
    PatchTST & 22.363 & \textbf{11.966} & 19.213 & 3.022 \\
    N-HiTS & \textbf{22.015} & 12.595 & 19.464 & 3.010 \\
    DLinear & 22.428 & 15.192 & \textbf{18.775} & \textbf{2.994} \\
    TiDE & 24.195 & 13.831 & 19.773 & 3.121 \\
    \bottomrule
  \end{tabular}
  \caption{sMAPE (\%) for each model and dataset. All forecasting models are trained once and not retrained across different seeds, ensuring that uncertainty estimates reflect only the variability of the counterfactual methods and are not confounded by forecasting model randomness. Consequently, results are reported for a single seed (seed 0) for the forecasting models, while the CF Models results are summarized across all tested seeds.}
  \label{appendix:tab_smape_table}
\end{table}

\subsection{Trade-off Between Proximity and Validity}
\label{appendix:proximity:tradeoff}

\begin{table}[!htbp]
\centering
\small
\begin{tabular}{r r r r r r r}
\hline
Sparsity & Proximity & Type & Comp. $\uparrow$ & Prox. $\downarrow$ & S-AUC $\uparrow$ & Ratio $\uparrow$ \\
\hline
\hline
0.02 & 0.4 & MAE & 0.447 $\pm$ 0.077 & 15.977 $\pm$ 0.300 & 0.921 $\pm$ 0.062 & 0.971 $\pm$ 0.026 \\
0.01 & 0.6 & MAE & 0.553 $\pm$ 0.042 & 16.051 $\pm$ 0.172 & 0.827 $\pm$ 0.084 & 0.938 $\pm$ 0.033 \\
0.005 & 1.0 & MAE & 0.626 $\pm$ 0.014 & 14.765 $\pm$ 0.182 & 0.683 $\pm$ 0.012 & 0.818 $\pm$ 0.010 \\
0.0005 & 1.2 & MAE & 0.708 $\pm$ 0.031 & 13.231 $\pm$ 0.574 & 0.540 $\pm$ 0.046 & 0.738 $\pm$ 0.015 \\
0.0001 & 1.4 & MAE & 0.764 $\pm$ 0.017 & 11.812 $\pm$ 1.350 & 0.439 $\pm$ 0.073 & 0.664 $\pm$ 0.032 \\
0.0001 & 1.4 & MSE & 0.067 $\pm$ 0.001 & 5.894 $\pm$ 0.039 & 0.336 $\pm$ 0.005 & 0.603 $\pm$ 0.003 \\
\hline
\hline
\end{tabular}
\caption{
Trade-off between sparsity, proximity, and validity in ConTex.
While our primary objective prioritizes sparse and valid interventions,
we observe that achieving higher validity typically comes at the cost of increased proximity error.
This trade-off can be explicitly controlled through the choice of proximity loss
and the relative weighting of sparsity and proximity terms.
In particular, switching from MAE to MSE biases the model towards minimizing proximity,
resulting in denser interventions and reduced validity.
Overall, this demonstrates that ConTex does not enforce a fixed objective,
but instead enables flexible adaptation to different optimization priorities.
}
\label{tab:multi_seed_results}
\end{table}

\subsection{Tourism Ablation}

\begin{table}[!htbp]
  \centering
  \label{tab:ablation_tourism_nhits}
  \begin{tabular}{lcccc}
  \toprule
Variant & Ratio $\uparrow$ & S-AUC $\uparrow$ & Comp. $\uparrow$ & Prox. $\downarrow$ \\
  \midrule
ConTex Full & 0.764 $\pm$ 0.048 & 0.656 $\pm$ 0.065 & 0.517 $\pm$ 0.055 & 11.067 $\pm$ 0.640 \\
w/o $\mathcal{L}_{L1}$ (MSE) & 0.632 $\pm$ 0.013 & 0.579 $\pm$ 0.018 & 0.280 $\pm$ 0.050 & 5.411 $\pm$ 0.121 \\
w/o Cond. Encoder & 0.138 $\pm$ 0.012 & 0.085 $\pm$ 0.004 & 0.930 $\pm$ 0.026 & 1.464 $\pm$ 0.599 \\
w/o Decomposition & 0.666 $\pm$ 0.029 & 0.608 $\pm$ 0.012 & 0.209 $\pm$ 0.167 & 6.240 $\pm$ 1.105 \\
w/o Current Forecast & 0.621 $\pm$ 0.059 & 0.558 $\pm$ 0.052 & 0.373 $\pm$ 0.125 & 5.926 $\pm$ 1.231 \\
  \bottomrule
  \end{tabular}
  \caption{Ablation study on the Tourism dataset using N-HiTS to validate the NN5 findings. Values are reported as mean $\pm$ standard deviation across random seeds. Similar to NN5, also on Tourism, the full ConTex configuration reaches the best trade-off between validity and compactness, and in contrast to NN5, the validity is not improved by any ablation.}
  \label{tab:ablation_tourism_nhits}
\end{table}

\bibliographystyle{plainnat}
\newpage
\bibliography{references}

\begin{thebibliography}{33}
\providecommand{\natexlab}[1]{#1}
\providecommand{\url}[1]{\texttt{#1}}
\expandafter\ifx\csname urlstyle\endcsname\relax
  \providecommand{\doi}[1]{doi: #1}\else
  \providecommand{\doi}{doi: \begingroup \urlstyle{rm}\Url}\fi

\bibitem[Amara-Ouali et~al.(2023)Amara-Ouali, Fasiolo, Goude, and Yan]{AMARAOUALI20231272}
Yvenn Amara-Ouali, Matteo Fasiolo, Yannig Goude, and Hui Yan.
\newblock Daily peak electrical load forecasting with a multi-resolution approach.
\newblock \emph{International Journal of Forecasting}, 39\penalty0 (3):\penalty0 1272--1286, 2023.
\newblock ISSN 0169-2070.
\newblock \doi{https://doi.org/10.1016/j.ijforecast.2022.06.001}.
\newblock URL \url{https://www.sciencedirect.com/science/article/pii/S0169207022000929}.

\bibitem[Andrawis et~al.(2011)Andrawis, Atiya, and El-Shishiny]{ANDRAWIS2011672}
Robert~R. Andrawis, Amir~F. Atiya, and Hisham El-Shishiny.
\newblock Forecast combinations of computational intelligence and linear models for the nn5 time series forecasting competition.
\newblock \emph{International Journal of Forecasting}, 27\penalty0 (3):\penalty0 672--688, 2011.
\newblock ISSN 0169-2070.
\newblock \doi{https://doi.org/10.1016/j.ijforecast.2010.09.005}.
\newblock URL \url{https://www.sciencedirect.com/science/article/pii/S0169207010001445}.
\newblock Special Section 1: Forecasting with Artificial Neural Networks and Computational Intelligence Special Section 2: Tourism Forecasting.

\bibitem[Ateş et~al.(2021)Ateş, Aksar, Leung, and Coskun]{inproceedings111}
Emre Ateş, Burak Aksar, Vitus Leung, and Ayse Coskun.
\newblock Counterfactual explanations for multivariate time series.
\newblock pages 1--8, 05 2021.
\newblock \doi{10.1109/ICAPAI49758.2021.9462056}.

\bibitem[Bahri et~al.(2022)Bahri, Boubrahimi, and Hamdi]{bahri2022shapeletbasedcounterfactualexplanationsmultivariate}
Omar Bahri, Soukaina~Filali Boubrahimi, and Shah~Muhammad Hamdi.
\newblock Shapelet-based counterfactual explanations for multivariate time series, 2022.
\newblock URL \url{https://arxiv.org/abs/2208.10462}.

\bibitem[Bai et~al.(2018)Bai, Kolter, and Koltun]{bai2018empiricalevaluationgenericconvolutional}
Shaojie Bai, J.~Zico Kolter, and Vladlen Koltun.
\newblock An empirical evaluation of generic convolutional and recurrent networks for sequence modeling, 2018.
\newblock URL \url{https://arxiv.org/abs/1803.01271}.

\bibitem[Bento et~al.(2021)Bento, Saleiro, Cruz, Figueiredo, and Bizarro]{Bento_2021}
João Bento, Pedro Saleiro, André~F. Cruz, Mário~A.T. Figueiredo, and Pedro Bizarro.
\newblock Timeshap: Explaining recurrent models through sequence perturbations.
\newblock In \emph{Proceedings of the 27th ACM SIGKDD Conference on Knowledge Discovery \& Data Mining}, KDD ’21, page 2565–2573. ACM, August 2021.
\newblock \doi{10.1145/3447548.3467166}.
\newblock URL \url{http://dx.doi.org/10.1145/3447548.3467166}.

\bibitem[Challu et~al.(2022)Challu, Olivares, Oreshkin, Garza, Mergenthaler-Canseco, and Dubrawski]{challu2022nhitsneuralhierarchicalinterpolation}
Cristian Challu, Kin~G. Olivares, Boris~N. Oreshkin, Federico Garza, Max Mergenthaler-Canseco, and Artur Dubrawski.
\newblock N-hits: Neural hierarchical interpolation for time series forecasting, 2022.
\newblock URL \url{https://arxiv.org/abs/2201.12886}.

\bibitem[Dai et~al.(2021)Dai, Meng, Dai, Wang, and Chen]{dai2021electricalpeakdemandforecasting}
Shuang Dai, Fanlin Meng, Hongsheng Dai, Qian Wang, and Xizhong Chen.
\newblock Electrical peak demand forecasting- a review, 2021.
\newblock URL \url{https://arxiv.org/abs/2108.01393}.

\bibitem[Das et~al.(2024)Das, Kong, Leach, Mathur, Sen, and Yu]{das2024longtermforecastingtidetimeseries}
Abhimanyu Das, Weihao Kong, Andrew Leach, Shaan Mathur, Rajat Sen, and Rose Yu.
\newblock Long-term forecasting with tide: Time-series dense encoder, 2024.
\newblock URL \url{https://arxiv.org/abs/2304.08424}.

\bibitem[Delaney et~al.(2021)Delaney, Greene, and Keane]{delaney2021instancebasedcounterfactualexplanationstime}
Eoin Delaney, Derek Greene, and Mark~T. Keane.
\newblock Instance-based counterfactual explanations for time series classification, 2021.
\newblock URL \url{https://arxiv.org/abs/2009.13211}.

\bibitem[Ferchichi et~al.(2025)Ferchichi, Abbes, Barra, and Farah]{ferchichi_trustworthy_2025}
A.~Ferchichi, A.B. Abbes, V.~Barra, and I.R. Farah.
\newblock Trustworthy {AI} for {Spatio}-{Temporal} {Forecasting} via {Counterfactual} {Causality}.
\newblock pages 10--17, 2025.
\newblock \doi{10.1109/ICTAI66417.2025.00010}.
\newblock URL \url{https://www.scopus.com/inward/record.uri?eid=2-s2.0-105031887993&doi=10.1109%2FICTAI66417.2025.00010&partnerID=40&md5=d6de56c36f3abd80813868f0f54b221b}.

\bibitem[Godahewa et~al.(2020{\natexlab{a}})Godahewa, Bergmeir, Webb, Hyndman, and Montero-Manso]{godahewa_electricity_2020}
Rakshitha Godahewa, Christoph Bergmeir, Geoff Webb, Rob Hyndman, and Pablo Montero-Manso.
\newblock Electricity {Hourly} {Dataset}, June 2020{\natexlab{a}}.
\newblock URL \url{https://zenodo.org/record/3889829}.

\bibitem[Godahewa et~al.(2020{\natexlab{b}})Godahewa, Bergmeir, Webb, Hyndman, and Montero-Manso]{godahewa_tourism_2020}
Rakshitha Godahewa, Christoph Bergmeir, Geoff Webb, Rob Hyndman, and Pablo Montero-Manso.
\newblock Tourism {Monthly} {Dataset}, June 2020{\natexlab{b}}.
\newblock URL \url{https://zenodo.org/record/3889425}.

\bibitem[Hahn et~al.(2026)Hahn, Königsfeld, Tercan, and Meisen]{hahn2026excoderexplainableclassificationdiscrete}
Yannik Hahn, Antonin Königsfeld, Hasan Tercan, and Tobias Meisen.
\newblock Excoder: Explainable classification of discrete time series representations, 2026.
\newblock URL \url{https://arxiv.org/abs/2602.13087}.

\bibitem[Hochreiter and Schmidhuber(1997)]{10.1162/neco.1997.9.8.1735}
Sepp Hochreiter and J\"{u}rgen Schmidhuber.
\newblock Long short-term memory.
\newblock \emph{Neural Comput.}, 9\penalty0 (8):\penalty0 1735–1780, November 1997.
\newblock ISSN 0899-7667.
\newblock \doi{10.1162/neco.1997.9.8.1735}.
\newblock URL \url{https://doi.org/10.1162/neco.1997.9.8.1735}.

\bibitem[Kim et~al.(2025)Kim, Kim, Kim, Lee, and Yoon]{kim_comprehensive_2025}
Jongseon Kim, Hyungjoon Kim, HyunGi Kim, Dongjun Lee, and Sungroh Yoon.
\newblock A comprehensive survey of deep learning for time series forecasting: architectural diversity and open challenges.
\newblock \emph{Artificial Intelligence Review}, 58\penalty0 (7):\penalty0 216, April 2025.
\newblock ISSN 1573-7462.
\newblock \doi{10.1007/s10462-025-11223-9}.
\newblock URL \url{https://link.springer.com/10.1007/s10462-025-11223-9}.

\bibitem[Ko et~al.(2023)Ko, Kim, Jeon, Ji, Chung, Suh, Chung, and Cho]{ko_deep_2023}
Ryoung-Eun Ko, Zero Kim, Bomi Jeon, Migyeong Ji, Chi~Ryang Chung, Gee~Young Suh, Myung~Jin Chung, and Baek~Hwan Cho.
\newblock Deep {Learning}-{Based} {Early} {Warning} {Score} for {Predicting} {Clinical} {Deterioration} in {General} {Ward} {Cancer} {Patients}.
\newblock \emph{Cancers}, 15\penalty0 (21):\penalty0 5145, October 2023.
\newblock ISSN 2072-6694.
\newblock \doi{10.3390/cancers15215145}.
\newblock URL \url{https://www.mdpi.com/2072-6694/15/21/5145}.

\bibitem[Lim et~al.(2020)Lim, Arik, Loeff, and Pfister]{lim2020temporalfusiontransformersinterpretable}
Bryan Lim, Sercan~O. Arik, Nicolas Loeff, and Tomas Pfister.
\newblock Temporal fusion transformers for interpretable multi-horizon time series forecasting, 2020.
\newblock URL \url{https://arxiv.org/abs/1912.09363}.

\bibitem[Luo and Yin(2026)]{luo_counterfactual_2026}
X.~Luo and W.~Yin.
\newblock Counterfactual {Explanation}-{Based} {Cryptocurrency} {Price} {Prediction}.
\newblock \emph{Entropy}, 28\penalty0 (1), 2026.
\newblock \doi{10.3390/e28010065}.
\newblock URL \url{https://www.scopus.com/inward/record.uri?eid=2-s2.0-105028503537&doi=10.3390%2Fe28010065&partnerID=40&md5=76f11583c4b6477488a22a29f85fa8c4}.

\bibitem[Makridakis et~al.(2020)Makridakis, Spiliotis, and Assimakopoulos]{MAKRIDAKIS202054}
Spyros Makridakis, Evangelos Spiliotis, and Vassilios Assimakopoulos.
\newblock The m4 competition: 100,000 time series and 61 forecasting methods.
\newblock \emph{International Journal of Forecasting}, 36\penalty0 (1):\penalty0 54--74, 2020.
\newblock ISSN 0169-2070.
\newblock \doi{https://doi.org/10.1016/j.ijforecast.2019.04.014}.
\newblock URL \url{https://www.sciencedirect.com/science/article/pii/S0169207019301128}.
\newblock M4 Competition.

\bibitem[Nayebi et~al.(2023)Nayebi, Tipirneni, Reddy, Foreman, and Subbian]{nayebi2023windowshapefficientframeworkexplaining}
Amin Nayebi, Sindhu Tipirneni, Chandan~K Reddy, Brandon Foreman, and Vignesh Subbian.
\newblock Windowshap: An efficient framework for explaining time-series classifiers based on shapley values, 2023.
\newblock URL \url{https://arxiv.org/abs/2211.06507}.

\bibitem[Nguyen and Ifrim(2026)]{10.1007/978-3-032-06078-5_4}
Thach~Le Nguyen and Georgiana Ifrim.
\newblock Tshap: Fast and exact shap for explaining time series classification and regression.
\newblock In Rita~P. Ribeiro, Bernhard Pfahringer, Nathalie Japkowicz, Pedro Larra{\~{n}}aga, Al{\'i}pio~M. Jorge, Carlos Soares, Pedro~H. Abreu, and Jo{\~a}o Gama, editors, \emph{Machine Learning and Knowledge Discovery in Databases. Research Track}, pages 60--77, Cham, 2026. Springer Nature Switzerland.
\newblock ISBN 978-3-032-06078-5.

\bibitem[Nie et~al.(2023)Nie, Nguyen, Sinthong, and Kalagnanam]{nie2023timeseriesworth64}
Yuqi Nie, Nam~H. Nguyen, Phanwadee Sinthong, and Jayant Kalagnanam.
\newblock A time series is worth 64 words: Long-term forecasting with transformers, 2023.
\newblock URL \url{https://arxiv.org/abs/2211.14730}.

\bibitem[Perez et~al.(2017)Perez, Strub, de~Vries, Dumoulin, and Courville]{perez2017filmvisualreasoninggeneral}
Ethan Perez, Florian Strub, Harm de~Vries, Vincent Dumoulin, and Aaron Courville.
\newblock Film: Visual reasoning with a general conditioning layer, 2017.
\newblock URL \url{https://arxiv.org/abs/1709.07871}.

\bibitem[Salinas et~al.(2020)Salinas, Flunkert, Gasthaus, and Januschowski]{SALINAS20201181}
David Salinas, Valentin Flunkert, Jan Gasthaus, and Tim Januschowski.
\newblock Deepar: Probabilistic forecasting with autoregressive recurrent networks.
\newblock \emph{International Journal of Forecasting}, 36\penalty0 (3):\penalty0 1181--1191, 2020.
\newblock ISSN 0169-2070.
\newblock \doi{https://doi.org/10.1016/j.ijforecast.2019.07.001}.
\newblock URL \url{https://www.sciencedirect.com/science/article/pii/S0169207019301888}.

\bibitem[Schlegel and Seidl(2026)]{schlegel2026whatifexplanationstimecounterfactuals}
Udo Schlegel and Thomas Seidl.
\newblock What-if explanations over time: Counterfactuals for time series classification, 2026.
\newblock URL \url{https://arxiv.org/abs/2603.27792}.

\bibitem[Theissler et~al.(2022)Theissler, Spinnato, Schlegel, and Guidotti]{theissler2022explainable}
Andreas Theissler, Francesco Spinnato, Udo Schlegel, and Riccardo Guidotti.
\newblock Explainable ai for time series classification: a review, taxonomy and research directions.
\newblock \emph{Ieee Access}, 10:\penalty0 100700--100724, 2022.

\bibitem[Wachter et~al.(2018)Wachter, Mittelstadt, and Russell]{wachter2018counterfactualexplanationsopeningblack}
Sandra Wachter, Brent Mittelstadt, and Chris Russell.
\newblock Counterfactual explanations without opening the black box: Automated decisions and the gdpr, 2018.
\newblock URL \url{https://arxiv.org/abs/1711.00399}.

\bibitem[Wang et~al.(2023)Wang, Miliou, Samsten, and Papapetrou]{wang_counterfactual_2023}
Z.~Wang, I.~Miliou, I.~Samsten, and P.~Papapetrou.
\newblock Counterfactual {Explanations} for {Time} {Series} {Forecasting}.
\newblock pages 1391--1396, 2023.
\newblock \doi{10.1109/ICDM58522.2023.00180}.
\newblock URL \url{https://www.scopus.com/inward/record.uri?eid=2-s2.0-85185401353&doi=10.1109%2FICDM58522.2023.00180&partnerID=40&md5=206d3e362e2089be0f73b3aeca43966c}.

\bibitem[Wang et~al.(2024)Wang, Samsten, Miliou, and Papapetrou]{wang_comet_2024}
Z.~Wang, I.~Samsten, I.~Miliou, and P.~Papapetrou.
\newblock {COMET}: {Constrained} {Counterfactual} {Explanations} for {Patient} {Glucose} {Multivariate} {Forecasting}.
\newblock pages 502--507, 2024.
\newblock \doi{10.1109/CBMS61543.2024.00089}.
\newblock URL \url{https://www.scopus.com/inward/record.uri?eid=2-s2.0-85200437241&doi=10.1109%2FCBMS61543.2024.00089&partnerID=40&md5=a224e4dfcecfaf00ad2130609bd91307}.

\bibitem[Yan and Wang(2023)]{yan2023selfinterpretabletimeseriesprediction}
Jingquan Yan and Hao Wang.
\newblock Self-interpretable time series prediction with counterfactual explanations, 2023.
\newblock URL \url{https://arxiv.org/abs/2306.06024}.

\bibitem[Zeng et~al.(2022)Zeng, Chen, Zhang, and Xu]{zeng2022transformerseffectivetimeseries}
Ailing Zeng, Muxi Chen, Lei Zhang, and Qiang Xu.
\newblock Are transformers effective for time series forecasting?, 2022.
\newblock URL \url{https://arxiv.org/abs/2205.13504}.

\bibitem[Zuin and Veloso(2024)]{zuin_navigating_2024}
G.~Zuin and A.~Veloso.
\newblock Navigating {Time}'s {Possibilities}: {Plausible} {Counterfactual} {Explanations} for {Multivariate} {Time}-{Series} {Forecast} through {Genetic} {Algorithms}.
\newblock Number 2024, pages 2575--2582, 2024.
\newblock \doi{10.1109/TrustCom63139.2024.00359}.
\newblock URL \url{https://www.scopus.com/inward/record.uri?eid=2-s2.0-105006506164&doi=10.1109%2FTrustCom63139.2024.00359&partnerID=40&md5=6a1191858aa7346ef48d19159a09185f}.

\end{thebibliography}

\newpage

\end{document}